\documentclass[letterpaper]{article} 
\usepackage{aaai25}  
\usepackage{times}  
\usepackage{helvet}  
\usepackage{courier}  
\usepackage[hyphens]{url}  
\usepackage{graphicx} 
\usepackage{natbib}  
\usepackage{caption} 
\usepackage[colorlinks=true, linkcolor=red, citecolor=blue, urlcolor=blue, pdfborder={0 0 0}]{hyperref}

\usepackage{algorithm}
\usepackage{algorithmic}

\usepackage{amssymb}
\usepackage{amsmath}
\usepackage{bm}
\usepackage{booktabs}
\usepackage{xcolor}
\usepackage{colortbl}

\usepackage{cleveref}
\crefname{equation}{Eq.}{Eqs.}
\Crefname{equation}{Eq.}{Eqs.}

\newtheorem{thm}{Theorem}

\newtheorem{defi}{Definition}

\usepackage{newfloat}
\usepackage{listings}
\DeclareCaptionStyle{ruled}{labelfont=normalfont,labelsep=colon,strut=off} 
\floatstyle{ruled}
\newfloat{listing}{tb}{lst}{}
\floatname{listing}{Listing}
\title{Parametric Pareto Set Learning for Expensive Multi-Objective Optimization}
\author{
    Ji Cheng, 
    Bo Xue, 
    Qingfu Zhang
}
\affiliations{
    Department of Computer Science, City University of Hong Kong\\

    \{J.Cheng, boxue4-c\}@my.cityu.edu.hk, qingfu.zhang@cityu.edu.hk
}

\usepackage{etoolbox}
\newcommand{\appendicesmode}{false}   

\begin{document}
\nocopyright

\maketitle

\begin{abstract}

Parametric multi-objective optimization (PMO) addresses the challenge of solving an infinite family of multi-objective optimization problems, where optimal solutions must adapt to varying parameters. Traditional methods require re-execution for each parameter configuration, leading to prohibitive costs when objective evaluations are computationally expensive. To address this issue, we propose Parametric Pareto Set Learning with multi-objective Bayesian Optimization (PPSL-MOBO), a novel framework that learns a unified mapping from both preferences and parameters to Pareto-optimal solutions. PPSL-MOBO leverages a hypernetwork with Low-Rank Adaptation (LoRA) to efficiently capture parametric variations, while integrating Gaussian process surrogates and hypervolume-based acquisition to minimize expensive function evaluations. We demonstrate PPSL-MOBO's effectiveness on two challenging applications: multi-objective optimization with shared components, where certain design variables must be identical across solution families due to modular constraints, and dynamic multi-objective optimization, where objectives evolve over time. Unlike existing methods that cannot directly solve PMO problems in a unified manner, PPSL-MOBO learns a single model that generalizes across the entire parameter space. By enabling instant inference of Pareto sets for new parameter values without retraining, PPSL-MOBO provides an efficient solution for expensive PMO problems. 

\end{abstract}

\section{Introduction}\label{sec:introduction}

The ability to make optimal decisions in the face of uncertain or varying parameters is a cornerstone of modern optimization and engineering design~\cite{fiacco1976sensitivity, bank1982non}. \emph{Parametric multi-objective optimization} (PMO) seeks to understand how the set of optimal trade-offs, known as the Pareto set (PS), evolves as exogenous parameters change~\cite{milgrom1994monotone, weaver2020parametric}. This paradigm is central to adaptive systems in science and engineering, where it is essential not only to solve for the best trade-offs under given conditions, but also to efficiently track the evolution of these trade-offs as requirements, environments, or constraints shift~\cite{wang2000optimal, tsai2021methodology}.

In this paper, we consider the following PMO problem:
\begin{equation}\label{eq:pmo}
\mathcal{F}^*(\bm{t}) \subset \arg\min_{\bm{x}\in\mathcal{X}} \left( f_1(\bm{x},\bm{t}), f_2(\bm{x},\bm{t}), \ldots, f_m(\bm{x},\bm{t})\right),
\end{equation}
where $\bm{x}$ represents the decision variable vector in the decision space $\mathcal{X}\subset\mathbb{R}^n$, $\bm{t}$ denotes parameters in the space $\Theta\subset\mathbb{R}^p$, and $\mathcal{F}^*(\bm{t})$ defines the optimal solutions that simultaneously minimize the $m$ objectives. Applications of PMO span diverse domains: from engineering design where operating conditions vary, to personalized medicine where treatments must adapt to patient characteristics, to autonomous systems that must respond to changing environments in real-time.

However, solving PMO problems presents fundamental challenges, particularly in \emph{expensive} settings where each objective evaluation requires costly simulations, physical experiments, or time-consuming computations. The core challenges include: (i) \textbf{Computational intractability}: Traditional methods require re-solving the entire optimization problem for each new parameter value, making real-time adaptation impossible; (ii) \textbf{Sample efficiency}: With limited evaluation budgets, we must learn the PS across the entire parameter space without exhaustive sampling; (iii) \textbf{Generalization}: The learned model must accurately predict Pareto-optimal solutions for previously unseen parameter values; and (iv) \textbf{Exploration-exploitation balance}: We must intelligently allocate evaluations between improving known regions and exploring new parameter territories.

While recent advances in Pareto set learning (PSL) have shown promise in learning mappings from preferences to Pareto-optimal solutions for fixed problems~\cite{navon2020learning, lin2022paretonips}, these methods fail to address the parametric nature of many real-world scenarios. Similarly, existing multi-objective Bayesian optimization approaches~\cite{knowles2006parego, emmerich2006single} focus on single-instance problems and cannot leverage the structural relationships across different parameter values. The lack of methods that can efficiently learn and generalize across both preference and parameter spaces represents a critical gap in the literature.

This work proposes \emph{Parametric Pareto Set Learning with Multi-Objective Bayesian Optimization} (PPSL-MOBO), a novel framework that addresses these challenges through three key innovations. First, we develop a unified neural architecture that learns the continuous mapping from both preferences and parameters to Pareto-optimal solutions, enabling instant adaptation to new scenarios. Second, we integrate surrogate modeling with an information-theoretic acquisition strategy to maximize learning efficiency under tight evaluation budgets. Third, we introduce a systematic approach to balance exploration across the parameter space with exploitation of promising regions, ensuring robust performance across diverse problem instances.

Our main contributions are:
\begin{itemize}
    \item We proposed a unified model architecture that learns parametric PSs as a continuous function of both preferences and exogenous parameters, enabling real-time multi-objective decision making without retraining.
    \item We developed a sample-efficient learning framework that integrates flexible surrogate models with intelligent acquisition strategies, dramatically reducing the number of expensive evaluations required.
    \item We applied our PPSL-MOBO method to modular design scenarios, where certain decision variables are shared across families of solutions. By modeling the PS as a function of shared components, our method enables efficient, real-time design for modular and customizable systems, as validated on real-world-inspired benchmarks.
    \item We extended PPSL-MOBO to dynamic multi-objective optimization, demonstrating that our approach can efficiently adapt to temporal changes by treating time as a parametric context. Extensive experiments on established dynamic benchmarks verify significant improvements in effectiveness over state-of-the-art methods.
\end{itemize} 
\section{Preliminaries}\label{sec:preliminaries}

\subsection{Multi-Objective Optimization}

In the context of expensive continuous multi-objective optimization, the goal is to optimize multiple conflicting objectives simultaneously, which can be formally expressed as:
\begin{equation}
    \min_{\bm{x}\in\mathcal{X}}\ \bm{f}(\bm{x}) = \left(f_1(\bm{x}), f_2(\bm{x}),\ldots,f_m(\bm{x})\right),
\end{equation}
where $\bm{x}$ represents a solution in the decision space $\mathcal{X} \subseteq \mathbb{R}^n$. The objective function $\bm{f}: \mathcal{X} \to \mathbb{R}^m$ is vector-valued, with each $f_i(\bm{x})$ being an individual objective. Each objective $f_i(\bm{x})$ is assumed to be expensive to evaluate, meaning that function evaluations require significant computational resources or time. In practical scenarios, these objectives often conflict with each other, making it impossible to find a single solution that simultaneously optimizes all objectives.

\begin{defi}[Pareto Dominance]
    A solution $\bm{x}_a$ is said to dominate another solution $\bm{x}_b$, denoted as $\bm{x}_a \prec \bm{x}_b$, if and only if $f_i(\bm{x}_a)\leq f_i(\bm{x}_b)$ for $i\in\{1,\ldots,m\}$, and $\exists j\in\{1,\ldots,m\}$ such that $f_j(\bm{x}_a)< f_j(\bm{x}_b)$. Furthermore, if $f_i(\bm{x}_a) < f_i(\bm{x}_b), \forall i \in \{1, \dots, m\}$, $\bm{x}_a$ is said to strictly dominate $\bm{x}_b$, denoted by $\bm{x}_a \prec_{\mathrm{strict}} \bm{x}_b$.
\end{defi}

\begin{defi}[Pareto Optimality]
    A solution $\bm{x}^*$ is considered Pareto optimal if no other solution $\hat{\bm{x}} \in \mathcal{X}$ exists such that $\hat{\bm{x}} \prec \bm{x}^*$. Moreover, $\bm{x}^\prime$ is weakly Pareto optimal if there is no solution $\hat{\bm{x}} \prec_{\mathrm{strict}} \bm{x}^\prime$.
\end{defi}

\begin{defi}[Pareto Set and Pareto Front]
    The set of all Pareto optimal solutions is referred to as the Pareto set, denoted by $\mathcal{M}_{\mathrm{ps}} \subseteq \mathcal{X}$. The corresponding image of these solutions in the objective space, $\bm{f}(\mathcal{M}_{\mathrm{ps}}) = \left\{\bm{f}(\bm{x}) \mid \bm{x} \in \mathcal{M}_{\mathrm{ps}} \right\}$, is termed the Pareto front. Similarly, we define the weak Pareto set $\mathcal{M}_{\mathrm{weak}}$ and weakly Pareto front $\bm{f}(\mathcal{M}_{\mathrm{weak}})$.
\end{defi}

Under appropriate regularity conditions, both the PS and Pareto front (PF) typically form $(m-1)$-dimensional manifolds in their respective spaces \cite{hillermeier2001generalized, zhang2008rm}. Each solution in the PS represents a unique trade-off among the conflicting objectives, and the cardinality of the PS can be infinite for continuous optimization problems.

\subsection{Bayesian Optimization}
Due to the expensive nature of objective evaluations, traditional multi-objective optimization algorithms that require numerous function evaluations become impractical. Bayesian Optimization (BO) has emerged as a powerful paradigm for tackling such expensive optimization problems. By constructing surrogate models (typically Gaussian processes) from limited evaluation data and employing acquisition functions to guide the search process, BO can efficiently identify high-quality solutions while minimizing the number of expensive function evaluations required. This model-based approach is particularly well-suited for expensive multi-objective optimization, where each evaluation must be carefully selected to maximize the information gained about the underlying PS and front. We provide a brief introduction as follows, and interested readers can refer to \cite{garnett2023bayesian}.

Gaussian Process (GP) provides a probabilistic framework for modeling unknown functions in Bayesian optimization. For a single objective function, a GP is fully specified by its prior distribution over the function space:
\begin{equation}
f(\bm{x}) \sim \mathcal{GP}(\mu(\bm{x}), k(\bm{x}, \bm{x}')),
\end{equation}
where $\mu: \mathcal{X} \rightarrow \mathbb{R}$ denotes the mean function (often assumed to be zero), and $k: \mathcal{X} \times \mathcal{X} \rightarrow \mathbb{R}$ represents the covariance kernel function that encodes assumptions about the function's smoothness and behavior.

Given a dataset of $n$ evaluated solutions $\mathcal{D} = \{(\bm{x}^{(i)}, y^{(i)}) \mid i = 1, \ldots, n\}$ where $y^{(i)} = f(\bm{x}^{(i)})$, the GP posterior can be computed in closed form. For any new input $\bm{x}$, the posterior mean and variance are:

\begin{equation}
\begin{aligned}
\hat{\mu}(\bm{x}) &= \mu(\bm{x}) + \bm{k}^T \bm{K}^{-1} (\bm{y} - \bm{\mu}), \\
\hat{\sigma}^2(\bm{x}) &= k(\bm{x}, \bm{x}) - \bm{k}^T \bm{K}^{-1} \bm{k},
\end{aligned}
\end{equation}
where $\bm{k} = [k(\bm{x}, \bm{x}^{(1)}), \ldots, k(\bm{x}, \bm{x}^{(n)})]^\top$ is the covariance vector between the new point and observed data, $\bm{K}$ is the kernel matrix of observed points with $\bm{K}_{ij} = k(\bm{x}^{(i)}, \bm{x}^{(j)})$, $\bm{y} = [y^{(1)}, \ldots, y^{(n)}]^\top$ contains the observed function values, and $\bm{\mu} = [\mu(\bm{x}^{(1)}), \ldots, \mu(\bm{x}^{(n)})]^\top$.

For multi-objective optimization with $m$ objectives, independent GP models are typically constructed for each objective $f_i$, yielding posterior means $\hat{\bm{\mu}}(\bm{x}) = [\hat{\mu}_1(\bm{x}), \ldots, \hat{\mu}_m(\bm{x})]^\top$ and variances $\hat{\bm{\sigma}}^2(\bm{x}) = [\hat{\sigma}_1^2(\bm{x}), \ldots, \hat{\sigma}_m^2(\bm{x})]^\top$.

The predictive distribution provided by the Gaussian Process, specifically the posterior mean and variance, forms the foundation for intelligently selecting the next evaluation point in BO. The acquisition function $\alpha: \mathcal{X} \rightarrow \mathbb{R}$ serves as a utility measure that guides the selection of the next evaluation point by balancing exploration (sampling uncertain regions) and exploitation (sampling promising regions). The next evaluation point is chosen by optimizing the acquisition function:
\begin{equation}
\bm{x}^{(n+1)} = \arg\max_{\bm{x} \in \mathcal{X}} \alpha(\bm{x}; \mathcal{D}).
\end{equation}

In multi-objective settings, scalar acquisition functions must be extended to handle vector-valued objectives. A particularly effective approach is based on hypervolume improvement (HVI), which measures the increase in dominated volume when adding new solutions. Given a current PF approximation $\mathcal{Y}$ and a set of candidate solutions $\mathcal{X}_+ = \{\bm{x}^{(i)}\}_{i=1}^b$ with predicted objectives $\hat{\mathcal{Y}}_+ = \{\hat{\bm{f}}(\bm{x}) \mid \bm{x} \in \mathcal{X}_+\}$, the hypervolume improvement is:
\begin{equation}\label{eqn:hvi}
\mathrm{HVI}(\hat{\mathcal{Y}}_+, \mathcal{Y}) = \mathrm{HV}(\hat{\mathcal{Y}}_+ \cup \mathcal{Y}) - \text{HV}(\mathcal{Y}),
\end{equation}
where $\mathrm{HV}(\cdot)$ denotes the hypervolume indicator computed with respect to a reference point. The next batch of solutions is selected by maximizing this improvement:
\begin{equation}
\mathcal{X}_+ = {\arg\max}_{\mathcal{X}_+ \subset \mathcal{X}}\ \mathrm{HVI}(\hat{\mathcal{Y}}_+, \mathcal{Y}).
\end{equation}

\subsection{Pareto Set Learning}

Pareto Set Learning, also known as PSL \cite{lin2020controllable, navon2020learning, lin2022paretoiclr, lin2022paretonips}, offers a novel paradigm for multi-objective optimization by learning a single parametric model that represents the entire PS. Instead of searching for individual Pareto optimal solutions, PSL aims to capture the continuous mapping from preference vectors to corresponding optimal solutions.

The core idea of PSL is to model the PS as a parametric function:
\begin{equation}
\bm{x} = h_{\bm{\theta}}(\bm{\lambda}),
\end{equation}
where $\bm{\lambda} \in \Delta^{m-1}$ represents a preference vector on the $(m-1)$-simplex $\Delta^{m-1} = \{\bm{\lambda} \in \mathbb{R}^m_+ \mid \sum_{i=1}^m \lambda_i = 1\}$, and $h_{\bm{\theta}}: \Delta^{m-1} \rightarrow \mathcal{X}$ is a learnable model parameterized by $\bm{\theta}$. Each preference vector $\bm{\lambda}$ encodes a specific trade-off among objectives, and the model $h_{\bm{\theta}}$ maps these preferences to their corresponding Pareto optimal solutions.

The objective of PSL is to find optimal parameters $\bm{\theta}^*$ such that for any preference $\bm{\lambda}$, the output $\bm{x} = h_{\bm{\theta}^*}(\bm{\lambda})$ minimizes an aggregation function:
\begin{equation}
\min_{\bm{x} \in \mathcal{X}}\ l_{\text{agg}}(\bm{x} \mid \bm{\lambda}), \quad \forall \bm{\lambda} \in \Delta^{m-1},
\end{equation}
where $l_{\text{agg}}: \mathcal{X} \times \Delta^{m-1} \rightarrow \mathbb{R}$ is a scalarization function that combines multiple objectives into a single value based on the preference vector. This formulation naturally extends decomposition-based approaches \cite{zhang2007moea} by replacing finite populations with a continuous parametric representation.

Given a distribution $P_{\bm{\lambda}}$ over preferences, the PSL optimization problem becomes:
\begin{equation}
\min_{\bm{\theta}}\ \mathcal{L}(\bm{\theta}) = \mathbb{E}_{\bm{\lambda} \sim P_{\bm{\lambda}}} \left[ l_{\text{agg}}(h_{\bm{\theta}}(\bm{\lambda}) \mid \bm{\lambda}) \right].
\end{equation}

The choice of aggregation function $l_{\text{agg}}$ is crucial for the quality and coverage of the learned PS. The simplest approach is the weighted sum method, however, this linear scalarization can only identify solutions on the convex hull of the PF, missing important non-convex regions \cite{boyd2004convex, ehrgott2005multicriteria}.

To overcome this limitation, the Tchebycheff (TCH) scalarization provides superior theoretical guarantees \cite{miettinen1999nonlinear}:
\begin{equation}
\label{eq:tch_psl}
l_{\text{tch}}(\bm{x} \mid \bm{\lambda}) = \max_{i \in [m]} \left\{ \lambda_i (f_i(\bm{x}) - (z_i^* - \varepsilon)) \right\},
\end{equation}
where $z_i^* < \min_{\bm{x} \in \mathcal{X}} f_i(\bm{x})$ represents the ideal value for the $i$-th objective, and $\varepsilon > 0$ is a small constant ensuring numerical stability.

\begin{thm}[\citet{choo1983proper}]\label{thm:tch}
A feasible solution $\bm{x} \in \mathcal{X}$ is weakly Pareto optimal if and only if there exists a valid preference vector $\bm{\lambda} \in \Delta^{m-1}$ such that $\bm{x}$ is an optimal solution of the Tchebycheff scalarization (\ref{eq:tch_psl}).
\end{thm}

This theorem establishes that TCH scalarization can recover all weakly Pareto optimal solutions, making it a theoretically sound choice for PSL. Recently, the smooth Tchebycheff (STCH) aggregation \cite{lin2024smooth} has been proposed to address computational challenges of the non-smooth max operator:
\begin{equation}
l_{\text{stch}}(\bm{x} \mid \bm{\lambda}, \nu) = \nu \log \left( \sum_{j=1}^m e^{\left(\frac{\lambda_j (f_j(\bm{x}) - (z_j^* - \varepsilon))}{\nu}\right)} \right),
\end{equation}
where $\nu > 0$ is a smoothing parameter. For brevity, we denote $l_{\text{stch}}(\bm{x} \mid \bm{\lambda})$ when using a fixed smoothing parameter. This smooth approximation preserves the desirable properties of TCH while offering significant computational advantages, including improved convergence rates and reduced per-iteration complexity due to its differentiability.

The theoretical foundation of STCH is particularly strong, as it maintains the ability to recover the entire PS under appropriate conditions:

\begin{thm}[\citet{lin2024smooth}]\label{thm:stch}
There exists a smoothing parameter $\nu^*$ such that for any $0 < \nu < \nu^*$, every Pareto solution of the multi-objective optimization problem corresponds to an optimal solution of the STCH aggregation with some valid preference vector $\bm{\lambda} \in \Delta^{m-1}$.
\end{thm}

The smooth nature of STCH not only facilitates gradient-based optimization but also enhances numerical stability, making it particularly well-suited for integration with modern machine learning frameworks. In this work, we adopt STCH as our primary aggregation method, though the proposed framework is sufficiently general to accommodate other scalarization approaches. This flexibility ensures that our methodology can be adapted to various problem characteristics and computational constraints in expensive multi-objective optimization scenarios.

\section{Parametric Pareto Set Learning for MOBO}\label{sec:method}

\begin{figure*}
    \centering
    \includegraphics[width=0.99\linewidth]{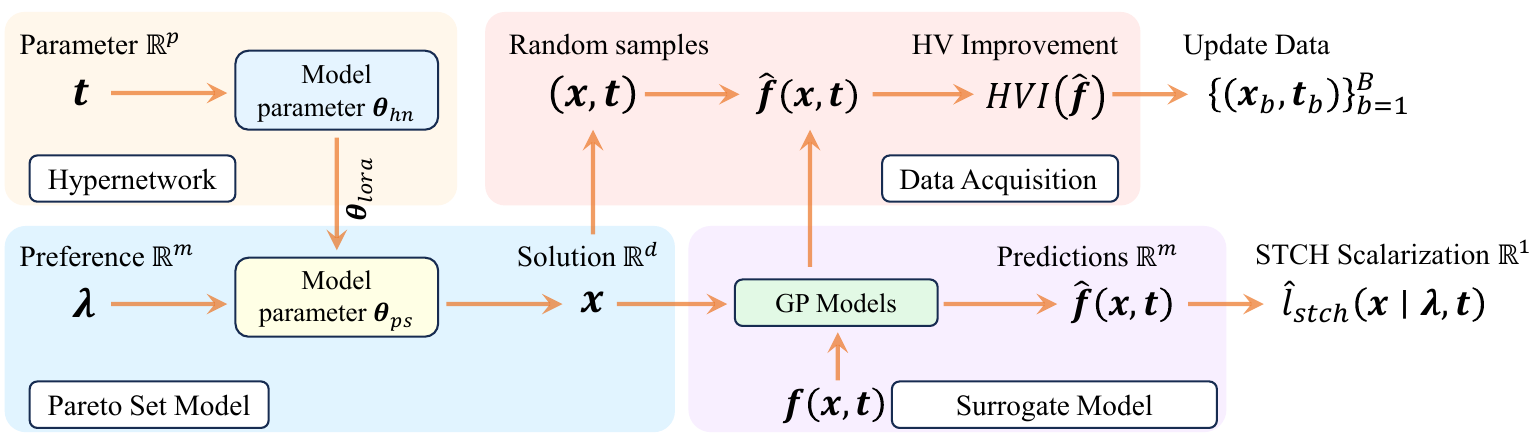}
    \caption{The PPSL-MOBO Framework. A hypernetwork adapts a PS model to a task parameter $\bm{t}$. The PS model generates approximate Pareto optimal solution $\bm{x}$ from preference $\bm{\lambda}$. The entire model is trained using predictions from surrogate models. A Bayesian optimization loop acquires new data by selecting PS-generated candidates that maximize HVI, which in turn refines the surrogate model.}
    \label{fig:ppsl_bo_paradigm}
\end{figure*}

The proposed PPSL-MOBO framework operates through three tightly coupled components: (1) the hypernetwork-LoRA architecture that generates task-specific PS models, (2) a Gaussian process-based training scheme that leverages surrogate models for scalable optimization, and (3) an intelligent data acquisition strategy that maximizes hypervolume improvement in the parametric space. As illustrated in Figure~\ref{fig:ppsl_bo_paradigm}, the system forms a closed-loop where newly acquired data continuously refines both the surrogate models and the parametric PS representation. The complete algorithmic workflow is formalized in Algorithm~1 (\Cref{app:subsec:algorithm_workflow}), which details the phased execution of surrogate updating, PS model training, and batch selection.

\subsection{Model Architecture}\label{subsec:model}

Our approach builds upon recent work in PSL \cite{lin2022paretonips}, which utilizes a neural network, $h_{\bm{\theta}_\mathrm{ps}}$, to model the mapping from a preference vector $\bm{\lambda}$ to a corresponding solution $\bm{x}$ on the PS:
\begin{equation}\label{eq:ps_model}
\bm{x} = h_{\bm{\theta}_\mathrm{ps}}(\bm{\lambda}).
\end{equation}
A key innovation of our work is to extend this model to handle parametric objectives, where the PS itself is a function of an external context or task parameter $\bm{t} \in \mathcal{T}$.

To achieve this, a straightforward approach would be to employ a hypernetwork, $g_{\bm{\theta}_\mathrm{hn}}$, that directly generates the entire set of weights $\bm{\theta}_\mathrm{ps}$ for the PS model based on the input parameter $\bm{t}$:
\begin{equation}\label{eq:hpn_before}
\bm{\theta}_\mathrm{ps} = g_{\bm{\theta}_\mathrm{hn}}(\bm{t}),
\end{equation}
where $\bm{\theta}_\mathrm{hn}$ are the learnable weights of the hypernetwork. However, this naive formulation presents a significant challenge. The task parameter $\bm{t}$ is typically low-dimensional, while the PS model's weights, $\bm{\theta}_\mathrm{ps}$, can be exceptionally high-dimensional (often numbering in the millions). This vast dimensionality gap makes the hypernetwork difficult to train, prone to overfitting, and unlikely to generalize well across the parameter space $\mathcal{T}$.

To overcome these challenges, we introduce a more sophisticated and parameter-efficient architecture based on Low-Rank Adaptation (LoRA). The central hypothesis is that the PSs for different task parameters $\bm{t}$ share a significant underlying structure. The LoRA framework elegantly captures this assumption by decomposing the PS model's weights into a large, shared base component and a small, task-specific, low-rank adaptation.

Formally, for each layer $l$ of the PS model, we express its weights $\bm{\theta}^l_\mathrm{ps}$ as the sum of shared base weights $\bm{\theta}^l_0$ and a low-rank matrix product $\bm{B}^l(\bm{t})\bm{A}^l(\bm{t})$:
\begin{equation}\label{eq:ps_theta}
\bm{\theta}^l_\mathrm{ps}(\bm{t}) = \bm{\theta}^l_0 + \bm{B}^l(\bm{t})\bm{A}^l(\bm{t}).
\end{equation}
Here, $\bm{\theta}^l_0$ is a trainable base parameter set shared across all tasks. The matrices $\bm{B}^l(\bm{t}) \in \mathbb{R}^{d^l \times r}$ and $\bm{A}^l(\bm{t}) \in \mathbb{R}^{r \times k^l}$ are the low-rank adapters, where the rank $r$ is a small integer (i.e., $r \ll \min(d^l, k^l)$).

Under this framework, the role of the hypernetwork is no longer to generate the full weights but only the compact low-rank matrices for all layers. Let $\bm{\theta}_\mathrm{lora}(\bm{t})$ represent the collection of all entries in $\bm{A}^l(\bm{t})$ and $\bm{B}^l(\bm{t})$ across all layers. The hypernetwork task is now redefined as:
\begin{equation}\label{eq:hn_lora}
\bm{\theta}_\mathrm{lora}(\bm{t}) = g_{\bm{\theta}_\mathrm{hn}}(\bm{t}).
\end{equation}
This approach dramatically reduces the output dimensionality of the hypernetwork. For a layer with original weight dimensions $d^l \times k^l$, the hypernetwork now only needs to predict $r(d^l + k^l)$ parameters instead of $d^l k^l$. This reduction not only enhances training efficiency and stability but also provides a powerful inductive bias, encouraging the model to learn a shared base structure of the PS manifold while capturing task-specific variations through low-rank modulations.

\subsection{Training Framework with Gaussian Process}\label{subsec:gp_train}

Our goal is to train the shared base weights $\bm{\theta}_0$ and the hypernetwork weights $\bm{\theta}_\mathrm{hn}$ such that the model's output, $\bm{x} = h_{\bm{\theta}_\mathrm{ps}(\bm{t})}(\bm{\lambda})$, is a Pareto-optimal solution for any given task parameter $\bm{t} \in \mathcal{T}$ and preference vector $\bm{\lambda} \in \Delta^{m-1}$.

Drawing upon the theoretical guarantees of the STCH scalarization (Theorems \ref{thm:tch} and \ref{thm:stch}), this goal can be formulated as finding the optimal solution to the following problem for all valid inputs:
\begin{equation}
\min_{\bm{x}\in\mathcal{X}}\ l_{\mathrm{stch}}(\bm{x} \mid \bm{t}, \bm{\lambda}),\quad \forall \bm{\lambda} \in \Delta^{m-1}, \bm{t}\in \mathcal{T}.
\end{equation}

To operationalize this, we frame the learning problem as minimizing the expected STCH loss over distributions of tasks and preferences. Let $P_{\bm{t}}$ be a distribution over the task parameter space $\mathcal{T}$, and let $P_{\bm{\lambda}}$ be a distribution over the preference simplex $\Delta^{m-1}$ (typically uniform to ensure full coverage). The global training objective is to minimize:
\begin{equation}\label{eq:train_obj}
\min_{\bm{\theta}_\mathrm{hn}, \bm{\theta}_0} \mathcal{L}(\bm{\theta}_\mathrm{hn}, \bm{\theta}_0) = \mathbb{E}_{\bm{t}\sim P_{\bm{t}},\bm{\lambda}\sim P_{\bm{\lambda}}} \left[l_\mathrm{stch}\left(h_{\bm{\theta}_\mathrm{ps}(\bm{t})}(\bm{\lambda}) \mid \bm{\lambda},\bm{t}\right)\right].
\end{equation}
However, in the context of expensive multi-objective optimization, the true objective functions $\bm{f}(\bm{x}; \bm{t})$ are not available for direct, large-scale gradient-based optimization. Therefore, we must rely on a surrogate model to approximate this loss function, which naturally leads us to integrate our PPSL framework with BO.

To make the surrogate models aware of the task parameter $\bm{t}$, we adopt a direct and effective input augmentation strategy. We define an augmented input space $\mathcal{Z} = \mathcal{X} \times \mathcal{T}$, where each point $\bm{z}$ is a concatenation of the decision variable $\bm{x}$ and the task parameter $\bm{t}$, i.e., $\bm{z} = [\bm{x}, \bm{t}]$. 
We then construct an independent GP model for each of the $m$ objective functions over this augmented space. For each objective $f_i$, the GP is defined as:
\begin{equation}
f_i(\bm{z}) \sim \mathcal{GP}(\mu_i(\bm{z}), k_i(\bm{z}, \bm{z}')).
\end{equation}
This approach allows the kernel function $k_i$ to directly model the influence of the task parameter $\bm{t}$ on the objective function, as well as any interaction effects between $\bm{x}$ and $\bm{t}$. The GP model can learn different lengthscales for the components of $\bm{x}$ and $\bm{t}$, effectively learning how sensitive the objective function is to changes in each part of the input.

The GP hyperparameters (lengthscales, variance) are optimized by maximizing the marginal log-likelihood on the full set of observed data $\mathcal{D} = \{(\bm{x}_j, \bm{t}_j, \bm{y}_j)\}_{j=1}^N$. Given a new augmented point $\bm{z} = [\bm{x}, \bm{t}]$, the posterior mean $\hat{\mu}_i(\bm{z})$ and variance $\hat{\sigma}_i^2(\bm{z})$ are computed in the standard way.

With the augmented-space GP models in place, we can define a tractable surrogate for the learning objective. To balance exploitation and exploration, we use the Lower Confidence Bound (LCB) as the acquisition signal. The surrogate objective vector $\hat{\bm{f}}$ for a given pair $(\bm{x}, \bm{t})$ is defined as:
\begin{equation}\label{eq:lcb_revised}
\hat{\bm{f}}(\bm{x}; \bm{t}) = \hat{\bm{\mu}}(\bm{z}) - \beta \hat{\bm{\sigma}}(\bm{z}),
\end{equation}
where $\beta$ is a hyperparameter controlling the exploration-exploitation trade-off, and $\hat{\bm{\mu}}$ and $\hat{\bm{\sigma}}$ are the posterior mean and standard deviation from our GPs, respectively.

By substituting the true objectives with this LCB acquisition signal, we arrive at the final, differentiable surrogate loss function that we can optimize via gradient descent:
\begin{equation}\label{eq:surrogate_loss_revised}
\hat{\mathcal{L}}(\bm{\theta}_{\text{hn}}, \bm{\theta}_{0}) = \mathbb{E}_{\bm{t}\sim P_{\bm{t}},\bm{\lambda}\sim P_{\bm{\lambda}}} \left[\hat{l}_{\text{stch}}\left(h_{\bm{\theta}_\mathrm{ps}(\bm{t})}(\bm{\lambda}) \mid \bm{\lambda},\bm{t}\right)\right],
\end{equation}
where $\hat{l}_{\text{stch}}$ is the surrogate STCH loss for a given task $\bm{t}$ and preference $\bm{\lambda}$, defined as:
\begin{equation}\label{eq:surrogate_stch_def}
\hat{l}_{\text{stch}}\left(\bm{x} \mid \bm{\lambda},\bm{t}\right) = \nu \log \left( \sum_{j=1}^m e^{\left(\frac{\lambda_j (\hat{f}_j(\bm{x}; \bm{t}) - (z_j^* - \varepsilon))}{\nu}\right)} \right).
\end{equation}
The expectation in \Cref{eq:surrogate_loss_revised} is approximated using Monte Carlo sampling. The gradients are backpropagated through the surrogate STCH loss, the LCB computation, and the parametric PS model to update the shared base weights $\bm{\theta}_{0}$ and the PS hypernetwork weights $\bm{\theta}_{\text{hn}}$. The details for the approximation, backpropagation, and the update of the parameters are presented in \Cref{subsec:monte_carlo} and \Cref{subsec:model_update}.

\subsection{Intelligent Data Acquisition in Parametric Space}\label{subsec:acq}

The final component of our PPSL-MOBO framework is the data acquisition strategy. At each iteration, we select a small batch of new task-solution pairs $(\bm{x}, \bm{t})$ to evaluate.

\textit{Generating a Candidate Pool from the Parametric Manifold.} 
First, we leverage our trained parametric PS model, $h_{\bm{\theta}_{\text{ps}}(\bm{t})}(\bm{\lambda})$, to generate a large pool of $P$ high-quality candidate points. We sample task-preference pairs $\{(\bm{t}_p, \bm{\lambda}_p)\}_{p=1}^P$ and generate corresponding solutions $\bm{x}_p = h_{\bm{\theta}_{\text{ps}}(\bm{t}_p)}(\bm{\lambda}_p)$, forming a candidate pool $\mathcal{C} = \{(\bm{x}_p, \bm{t}_p)\}_{p=1}^P$.

\textit{Batched Selection via Hypervolume Improvement.} 
From the candidate pool $\mathcal{C}$, we select a batch of $B$ points for expensive evaluation using a sequential greedy strategy based on HVI (\Cref{eqn:hvi}). We iteratively select the point from the candidate pool that offers the largest marginal HVI with respect to the current batch and the archive of already evaluated points $\mathcal{D}$. The HVI is calculated using the LCB value (\Cref{eq:lcb_revised}) from our augmented-space GP surrogates.

\section{Application Studies: Multi-Objective Optimization with Shared Components}\label{sec:shared_components}

In this section, we adapt our proposed PPSL-MOBO framework to a challenging and practical multi-objective design problem: optimization with shared components, where certain design variables are constrained to be identical across a family of solutions. The detailed setup and results can be found in \Cref{app:sec:experiment_mopsc}. 

\begin{table*}[h]
\scriptsize
\centering
\begin{tabular}{lc|ccccc}
\toprule[1.0pt]
Problem & Shared Comp. & NSGA-II & qParEGO & qEHVI & PSL-MOBO & PPSL-MOBO \\
\midrule\midrule
     RE21 & $(x_1)$ & 6.52e-01 (3.57e-03) & 6.96e-01 (2.89e-02) & 7.32e-01 (1.70e-04) & \cellcolor{gray!30}\textbf{7.34e-01} (5.79e-05) & 7.33e-01 (2.31e-03) \\
     & $(x_2)$ & 6.90e-01 (1.04e-02) & 7.53e-01 (5.45e-03) & 7.82e-01 (5.59e-05) & \cellcolor{gray!30}\textbf{7.84e-01} (3.43e-05) & 7.79e-01 (7.91e-03) \\
     & $(x_3)$ & 6.14e-01 (2.74e-03) & 6.53e-01 (6.11e-03) & 6.72e-01 (1.75e-04) & \cellcolor{gray!30}\textbf{6.76e-01} (8.41e-05) & 6.68e-01 (7.07e-03) \\
     & $(x_4)$ & 7.02e-01 (2.24e-03) & 7.70e-01 (1.15e-02) & 7.99e-01 (2.22e-04) & \cellcolor{gray!30}\textbf{8.02e-01} (4.01e-05) & 7.99e-01 (2.74e-03) \\
     & $(x_1,x_2)$ & 5.92e-01 (5.35e-03) & 6.19e-01 (1.77e-03) & \cellcolor{gray!30}\textbf{6.23e-01} (1.05e-05) & \cellcolor{gray!30}\textbf{6.23e-01} (1.64e-05) & \cellcolor{gray!30}\textbf{6.23e-01} (7.84e-04) \\
     & $(x_2,x_3)$ & 5.38e-01 (1.38e-03) & 5.58e-01 (4.16e-03) & 5.66e-01 (3.53e-05) & 5.67e-01 (7.38e-05) & \cellcolor{gray!30}\textbf{5.69e-01} (6.26e-05) \\
     & $(x_3,x_4)$ & 5.36e-01 (1.81e-03) & 5.60e-01 (9.90e-04) & 5.64e-01 (2.06e-05) & 5.65e-01 (1.86e-04) & \cellcolor{gray!30}\textbf{5.66e-01} (7.15e-04) \\
     & $(x_1,x_2,x_3)$ & 3.96e-01 (1.52e-04) & 3.98e-01 (1.21e-03) & \cellcolor{gray!30}\textbf{4.00e-01} (5.39e-06) & 3.99e-01 (7.76e-05) & \cellcolor{gray!30}\textbf{4.00e-01} (6.61e-06) \\
     & $(x_2,x_3,x_4)$ & 5.11e-01 (1.43e-03) & 5.15e-01 (3.12e-03) & 5.18e-01 (1.40e-05) & 5.18e-01 (7.88e-05) & \cellcolor{gray!30}\textbf{5.19e-01} (2.19e-05) \\
\bottomrule[1.0pt]
\end{tabular}
\caption{HV values of PPSL-MOBO and baseline approaches on MOPs with shared components.All baseline methods were re-executed with a budget of 100 evaluations per task across the ten tested problems, resulting 1,000 evaluations for each share component configuration. In contrast, PPSL-MOBO was trained only once, using a total of 200 evaluations, and subsequently inferred the PS for each parameterized task.}
\label{tab:mopsc_compare_baseline}
\end{table*}

\subsection{Problem Introduction}

In many real-world applications, structural constraints are applied to design variables due to manufacturing limitations, modular design principles, or computational budgets \cite{Guha2024Compromising, Lin2025Dealing, Zhao2025Component}. For example, in personalized manufacturing, a key challenge is to generate customized products that cater to individual preferences while maintaining cost efficiency through shared components. Each customer's preference defines a specific trade-off, and the corresponding personalized design can be viewed as a Pareto optimal solution. To minimize manufacturing costs, it is essential that these designs share common modules, ensuring both flexibility and economic feasibility \cite{garcia2019modular}.

Formally, let $\bm{x}_{\boldsymbol{s}}$ denote the subvector of decision variables $\bm{x}$ that holds the shared components, where $\bm{s} \subset \{1, \ldots, n\}$. Let $\bm{x}_{\bm{p}}$ be the remaining variables to be optimized, with $\bm{p}=\{1, \ldots, n\} \setminus \bm{s}$. The MOP with shared components can be expressed as finding the PS for a fixed set of shared values $\bm{\beta}$:
\begin{equation}
\mathcal{F}(\bm{\beta}) \subset\min_{\bm{x}_{\bm{p}}}\ \bm{F} (\bm{x}_{\bm{p}} | \bm{x}_{\bm{s}}=\bm{\beta}).
\end{equation}
In this context, the shared component values $\bm{\beta}$ act as the task parameter of our parametric problem. The goal is to learn a model that can generate the entire PS of optimizable variables $\bm{x}_{\bm{p}}$ for any given shared component configuration $\bm{x}_{\bm{s}}$. Using PPSL-MOBO, the parametric PS model is denoted by:
\begin{equation}
\bm{x}_{\bm{p}} = h_{\bm{\theta}(\bm{x}_{\bm{s}})}(\bm{\lambda}),
\end{equation}
where $h(\cdot)$ is a map from the preference space $\bm{\Delta}^{m-1}$ to the decision space of the non-shared variables $\mathbb{R}^{|\bm{p}|}$. The parameters of this map, $\bm{\theta}(\bm{x}_{\bm{s}})$, are generated by the hypernetwork conditioned on the shared component values $\bm{x}_{\bm{s}}$.

This formulation is particularly useful in scenarios where the shared components $\bm{x}_{\bm{s}}$ are determined by external constraints, predefined modules, or customer choices, and the primary design task is to find the optimal configuration of $\bm{x}_{\bm{p}}$ around these fixed components.

\subsection{Results Analysis}

We evaluate PPSL-MOBO against MOEA and MOBO methods on a real-world multi-objective test problems suite \cite{tanabe2020easy, lin2022paretonips}. The experimental results in Tables \ref{tab:mopsc_compare_baseline} and \ref{tab:mopsc_runtime_compare} (see \Cref{app:mopsc_res}) demonstrate the effectiveness and efficiency of our proposed method. PPSL-MOBO achieves competitive hypervolume performance across different shared component configurations, often matching or outperforming baseline methods that are specifically optimized for individual tasks. Additional results on more real-world problems are presented in \Cref{tab:mopsc_compare_baseline_addi}.
The most significant advantage of PPSL-MOBO lies in its computational and sample efficiency. While baseline methods must be re-executed from scratch for each new shared component configuration, PPSL-MOBO requires only a one-time training phase followed by near-instantaneous inference (in milliseconds) for new parameter values. This approach achieves substantial reductions in both total function evaluations and computational time compared to baseline methods, making it highly practical for applications requiring frequent parameter variations or expensive function evaluations.

\begin{figure*}
    \centering
    \includegraphics[width=0.95\linewidth]{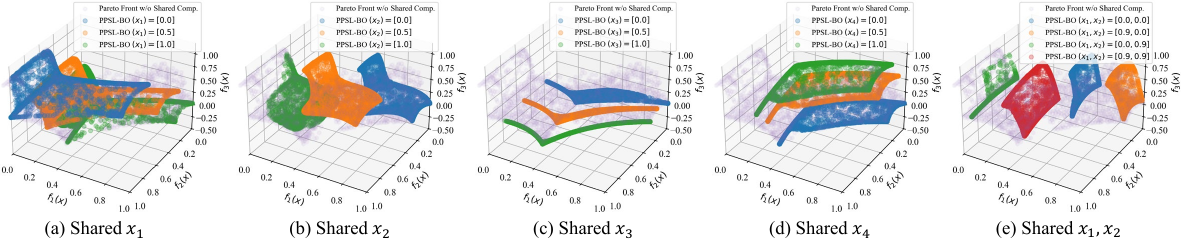}
    \caption{Learned PFs for the Rocket Injector Design Problem (RE37) with Shared Components. We demonstrate PFs for several specific parameter values, while PPSL-MOBO can obtain PS for any parameter in milliseconds. The transparent points denote the PFs for the problem without shared component. The shared components correspond to the decision variable $x_1$ (hydrogen flow angle), $x_2$ (hydrogen area), $x_3$ (oxygen area), and $x_4$ (oxidizer post tip thickness).}
    \label{fig:ps_re37}
\end{figure*}

Figure \ref{fig:ps_re37} illustrates the PFs learned for the Rocket Injector Design Problem (RE37) under different shared component configurations. The visualization demonstrates how constraining different design variables affects the shape and position of the achievable PFs. 
These visualizations confirm that PPSL-MOBO successfully captures the parametric variations in PF geometry across different shared component scenarios, providing decision-makers with comprehensive trade-off information for each design configuration.

\begin{figure*}
    \centering
    \includegraphics[width=0.95\linewidth]{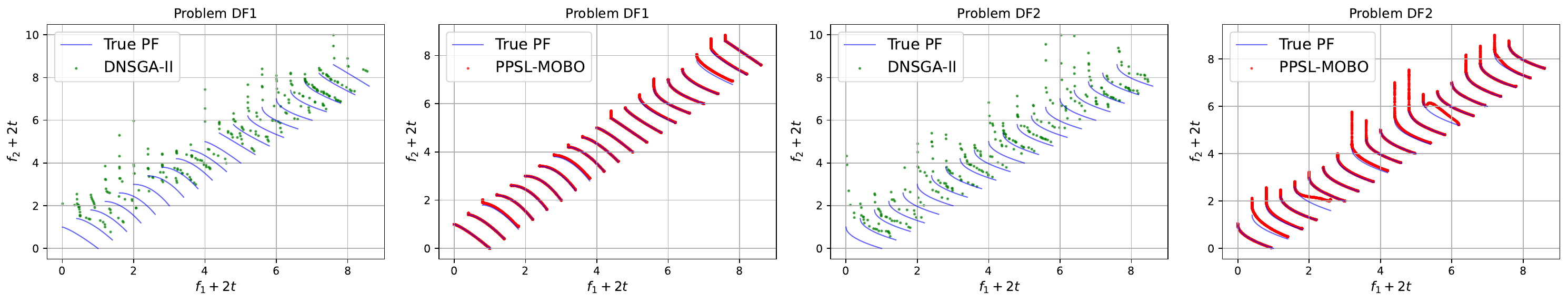}
    \caption{Comparison of generated solutions by DNSGA-II (green) and PPSL-MOBO (red) at each time step on DMOPs. 200 evaluations are allowed for each time step.}
    \label{fig:dmop_ps_scatter}
\end{figure*}

\section{Application Study: Dynamic Multi-Objective Optimization}\label{sec:dynamic_moo}

In this section, we demonstrate the effectiveness of the proposed PPSL-MOBO framework for Dynamic Multi-Objective Optimization Problems (DMOPs), where the optimization objectives change over time. 

\subsection{Problem Introduction}

Dynamic multi-objective optimization arises in numerous real-world scenarios, such as dynamic portfolio selection~\cite{xu2006power}, time-varying process control~\cite{tang2021data}, adaptive scheduling~\cite{abello2011adaptive}, and non-stationary machine learning~\cite{kim2010multiobjective}. In these settings, the parameter at \Cref{eq:pmo} becomes the time characteristic, inducing time-varying PS and PF to resolve sequentially. 

Traditional approaches to DMOPs often detect environmental changes and restart or re-initialize optimization at each change point, treating each time instance as an independent static problem \cite{zhou2013population, jiang2022evolutionary, zhang2023adaptive}. However, this disregards the inherent temporal correlation between solutions at adjacent time steps, potentially leading to inefficiency and suboptimal adaptation.
With PPSL-MOBO, we treat the temporal variable $t$ as a \emph{parametric context}, enabling the learning of a \emph{unified parametric PS model} that generalizes across the temporal dimension. This approach not only allows for efficient adaptation to environmental changes but also leverages knowledge transfer across time, capturing the underlying temporal dynamics of the evolving PS and PF.

To align with the algorithmic constraint that only the current time step's function evaluation is accessible, the task distribution $P_{\bm{t}}$ in \Cref{alg:PPSL-MOBO-final} is set to a degenerated distribution, yielding $N_t$ identical parameter samples at each generation. The experimental setup, implementation of PPSL-MOBO, and the compared methods are elaborated in \Cref{subsec:dmop_experimental_set}.

\subsection{Results and Analysis}



\Cref{fig:dmop_ps_scatter} illustrates the dynamic tracking capabilities of PPSL-MOBO compared to DNSGA-II \cite{deb2007dynamic} on benchmark problems DF1 and DF2 \cite{jiang2018benchmark}. The visualization highlights the fundamental differences in how these algorithms respond to environmental changes. DNSGA-II, despite being specifically designed for dynamic optimization, struggles to converge to the true Pareto front within the limited two-generation window between environmental changes. Its solutions (depicted as scattered points) remain widely dispersed and fail to adequately approximate the complete Pareto front structure. In contrast, PPSL-MOBO leverages its pre-trained parametric model to instantaneously generate well-distributed Pareto fronts (shown as continuous curves) that closely approximate the true Pareto front at each time step.

\Cref{tab:dmop_migd} presents a comprehensive comparison between PPSL-MOBO and state-of-the-art algorithms from the DMOP community across all DF benchmark problems. \Cref{fig:dmop_migd} and \ref{fig:dmop_mhv} visualize the performance trajectories throughout the dynamic optimization process. Finally, the ablation study results (\Cref{tab:dmop_migd_abla}) reveal the synergistic contributions of each component in PPSL-MOBO's architecture, validating our design choices for efficient dynamic optimization.
\section{Conclusion and Future Work}\label{sec:discussion_conclusions}

We presented PPSL-MOBO, a novel framework that learns parametric Pareto sets for expensive multi-objective optimization. By combining hypernetwork-LoRA architectures with Bayesian optimization, our approach efficiently maps both preferences and exogenous parameters to Pareto-optimal solutions through a single unified model, eliminating the need for retraining across parameter variations.

Our experiments demonstrate PPSL-MOBO's effectiveness in two challenging domains: (1) In multi-objective optimization with shared components, it achieves competitive performance using only a fraction of function evaluations compared to methods requiring re-execution for each parameter value. (2) In dynamic optimization, by treating time as a parametric context, PPSL-MOBO successfully adapts to changing environments and outperforms state-of-the-art dynamic MOEAs. These results validate that parametric learning of Pareto sets offers a powerful paradigm for expensive optimization scenarios.

Future work includes extending the framework to constrained multi-objective problems, developing theoretical guarantees on sample complexity, and exploring integration with interactive decision-making systems for real-time preference articulation.

\ifdefstring{\appendicesmode}{true}{%
}{%
  \bibliography{ref_all}%
}

\appendix
\onecolumn
\newpage

\section*{Appendix Overview}

This appendix provides supplementary materials to support the main paper. The content is organized as follows:
\begin{itemize}
\item \textbf{Appendix \ref{app:sec:algorithm}}: Provides \textbf{implementation details} of the PPSL-MOBO algorithm, including pseudocode for key subroutines, hyperparameter settings, and computational complexity analysis.
\item \textbf{Appendix \ref{app:sec:limitation_impact}}: Analyzes the \textbf{limitations} of our method and discusses its \textbf{potential societal impacts}, including ethical considerations and possible future directions.
\item \textbf{Appendix \ref{app:sec:related_work}}: Presents \textbf{detailed related work} relevant to this paper, including comprehensive comparisons with existing methods and discussions of their connections to our approach.
\item \textbf{Appendix \ref{app:sec:experiment_mopsc}}: Details the \textbf{multi-objective optimization with shared components} experiments, including experimental setup, baseline methods, and comprehensive results analysis.
\item \textbf{Appendix \ref{app:sec:experiment_dmop}}: Provides extensive coverage of the \textbf{dynamic multi-objective optimization} application, including experimental design, algorithmic adaptations of PPSL-MOBO, comparison methods, results analysis, and ablation studies.
\item \textbf{Appendix \ref{app:sec:license}}: Lists the \textbf{licenses} of all code dependencies used in this work and provides licensing information for our proposed PPSL-MOBO implementation.
\end{itemize}

\section{The PPSL-MOBO algorithm}\label{app:sec:algorithm}

\subsection{Algorithmic Workflow}\label{app:subsec:algorithm_workflow}
\begin{algorithm}[H]
\caption{PPSL-MOBO: Parametric Pareto Set Learning via Mult-Objective Bayesian Optimization}
\label{alg:PPSL-MOBO-final}
\begin{algorithmic}
    \STATE \textbf{Input:} PS model $h_{\bm{\theta}_{\text{ps}}}(\cdot)$, PS hypernetwork $g_{\bm{\theta}_\mathrm{hn}}(\cdot)$, Batch size $B$.
    \STATE \textbf{Initialize:} Collect an initial dataset $\mathcal{D} = \{(\bm{x}_j, \bm{t}_j, \bm{y}_j)\}_{j=1}^{N_0}$ using a space-filling design. Initialize model parameters $\bm{\theta}_0, \bm{\theta}_\mathrm{hn}$.
    \FOR{iteration $i = 1, 2, \ldots$}
    \STATE \textit{// Phase 1: Update Surrogate Models}
    \STATE Train $m$ independent GP models on the augmented inputs $[\bm{x}, \bm{t}]$ from the current dataset $\mathcal{D}$ by maximizing the marginal log-likelihood to fit the GP hyperparameters.
    \STATE \textit{// Phase 2: Update Parametric PS Model}
    \FOR{step $j = 1$ to $J$}
        \STATE Sample a batch of tasks $\{\bm{t}_k\}_{k=1}^{N_t} \sim P_{\bm{t}}$ and preferences $\{\bm{\lambda}_k\}_{k=1}^{N_{\lambda}} \sim P_{\bm{\lambda}}$.
        \STATE Compute the surrogate loss $\hat{\mathcal{L}}(\bm{\theta}_{\text{hn}}, \bm{\theta}_{0})$ using the LCB acquisition by \Cref{eq:surrogate_loss_revised}.
        \STATE Update PS model parameters $\bm{\theta}_0, \bm{\theta}_\mathrm{hn}$ using stochastic gradient descent: $\nabla_{\bm{\theta}_0, \bm{\theta}_\mathrm{hn}} \hat{\mathcal{L}}(\bm{\theta}_{\text{hn}}, \bm{\theta}_{0})$.
    \ENDFOR
    
    \STATE \textit{// Phase 3: Acquire New Data Batch via HVI}
    \STATE Sample $P \gg B$ task-preference pairs $\{(\bm{t}_p, \bm{\lambda}_p)\}_{p=1}^P$.
    \STATE Generate candidate solutions $\bm{x}_p = h_{\bm{\theta}_{\text{ps}}(\bm{t}_p)}(\bm{\lambda}_p)$, forming a pool $\mathcal{C} = \{(\bm{x}_p, \bm{t}_p)\}_{p=1}^P$.
    
    \STATE Initialize empty batch $\mathcal{B} \leftarrow \emptyset$.
    \FOR{$b = 1$ to $B$}
        \STATE Find $(\bm{x}^*, \bm{t}^*) = \arg\max_{(\bm{x}, \bm{t}) \in \mathcal{C} \setminus \mathcal{B}} \text{HVI}(\mathcal{B} \cup \{(\bm{x}, \bm{t})\} \mid \mathcal{D})$.
        \STATE Add $(\bm{x}^*, \bm{t}^*)$ to $\mathcal{B}$.
    \ENDFOR
    
    \STATE Evaluate the true objectives $\bm{Y}_{\mathcal{B}} = \{\bm{f}(\bm{x}_b; \bm{t}_b) \mid (\bm{x}_b, \bm{t}_b) \in \mathcal{B}\}$.
    \STATE Augment the dataset: $\mathcal{D} \leftarrow \mathcal{D} \cup \{(\bm{x}_b, \bm{t}_b, \bm{y}_b)\}_{(\bm{x}_b, \bm{t}_b) \in \mathcal{B}}$.
    \ENDFOR
    \STATE \textbf{Output:} Trained models $h$, $g_{\bm{\theta}_\mathrm{hn}}$, and final dataset $\mathcal{D}$.
\end{algorithmic}
\end{algorithm}

\subsection{Monte Carlo Approximation of the Training Objective.}\label{subsec:monte_carlo}

To optimize the parametric PS model in practice, we employ Monte Carlo sampling to approximate the expectation in the training objective (\Cref{eq:surrogate_loss_revised}). Specifically, we sample $N_t$ task parameters from the distribution $P_{\bm{t}}$ and, for each task parameter $\bm{t}_i$, we sample $N_{\lambda}$ preference vectors from the distribution $P_{\bm{\lambda}}$. The PPSL model then generates the corresponding solutions $\left\{h_{\bm{\theta}_\mathrm{ps}(\bm{t}_i)}(\bm{\lambda}_j)\right\}_{j=1}^{N_{\lambda}}$ for each sampled parameter-preference pair.

The surrogate training objective is thus approximated as:
\begin{equation}
\hat{\mathcal{L}}(\bm{\theta}_{\mathrm{hn}},\bm{\theta}_{0}) \approx \frac{1}{N_t}\sum_{i=1}^{N_t}\frac{1}{N_{\lambda}}\sum_{j=1}^{N_{\lambda}} \hat{l}_\mathrm{stch}\left(h_{\bm{\theta}_\mathrm{ps}(\bm{t}_i)}(\bm{\lambda}_j)\mid \bm{\lambda}_j, \bm{t}_i\right),
\end{equation}
where $\hat{l}_\mathrm{stch}$ is the surrogate STCH loss defined in \Cref{eq:surrogate_stch_def}, which uses the LCB predictions from the GP models rather than true objective evaluations.

Under appropriate regularity conditions on the surrogate STCH loss function, we can interchange the expectation and differentiation operators to compute gradients with respect to the model parameters.

The gradient of the surrogate loss with respect to the shared base parameters $\bm{\theta}_0$ is:
\begin{equation}
\begin{aligned}
\nabla_{\bm{\theta}_{0}} \hat{\mathcal{L}} & = \nabla_{\bm{\theta}_{0}} \mathbb{E}_{\bm{t}\sim P_{\bm{t}},\bm{\lambda}\sim P_{\bm{\lambda}}} \left[\hat{l}_\mathrm{stch}\left(h_{\bm{\theta}_\mathrm{ps}(\bm{t})}(\bm{\lambda})\mid \bm{\lambda}, \bm{t}\right)\right] \\
& = \mathbb{E}_{\bm{t}\sim P_{\bm{t}},\bm{\lambda}\sim P_{\bm{\lambda}}}  \left[\nabla_{\bm{\theta}_{0}}\hat{l}_\mathrm{stch}\left(h_{\bm{\theta}_\mathrm{ps}(\bm{t})}(\bm{\lambda})\mid \bm{\lambda}, \bm{t}\right)\right].
\end{aligned}
\end{equation}

Using the Monte Carlo approximation:
\begin{equation}
\nabla_{\bm{\theta}_{0}} \hat{\mathcal{L}} \approx \frac{1}{N_tN_{\lambda}}\sum_{i=1}^{N_t}\sum_{j=1}^{N_{\lambda}} \nabla_{\bm{\theta}_{0}} \hat{l}_\mathrm{stch}\left(h_{\bm{\theta}_\mathrm{ps}(\bm{t}_i)}(\bm{\lambda}_j)\mid \bm{\lambda}_j, \bm{t}_i\right).
\end{equation}

For each sample, the gradient is computed via the chain rule:
\begin{equation}\label{eq:gra_theta0_app}
\nabla_{\bm{\theta}_{0}} \hat{l}_\mathrm{stch}=\frac{\partial \bm{\theta}_\mathrm{ps}}{\partial \bm{\theta}_0}\cdot \frac{\partial h_{\bm{\theta}_\mathrm{ps}}(\bm{\lambda})}{\partial \bm{\theta}_\mathrm{ps}}\cdot \nabla_{\bm{x}} \hat{l}_\mathrm{stch}(\bm{x}\mid\bm{\lambda},\bm{t}),
\end{equation}
where $\frac{\partial \bm{\theta}_\mathrm{ps}}{\partial \bm{\theta}_0} = \mathbf{I}$ (identity matrix) from \Cref{eq:ps_theta}, since $\bm{\theta}_0$ appears directly in the parametrization.

To compute $\nabla_{\bm{x}} \hat{l}_\mathrm{stch}(\bm{x}\mid\bm{\lambda},\bm{t})$, we follow the derivation for the standard STCH scalarization but substitute the true objectives with the LCB predictions from our GP models.

Let $\bm{y} = \bm{\lambda} \odot (\hat{\bm{f}}(\bm{x}, \bm{t}) - \bm{z}^*)$ be an $m$-dimensional vector, where $\odot$ denotes element-wise multiplication and $\hat{\bm{f}}(\bm{x}; \bm{t})$ is the LCB prediction from \Cref{eq:lcb_revised}. The surrogate STCH loss can be rewritten as:
\begin{equation}
\hat{l}_\mathrm{stch}(\bm{x}\mid\bm{\lambda},\bm{t}) = \nu \log \left( \sum_{i=1}^m e^{y_i/\nu} \right).
\end{equation}

Taking the gradient with respect to $\bm{y}$:
\begin{equation}
\nabla_{\bm{y}} \hat{l}_\mathrm{stch} = \nabla_{\bm{y}} \nu \log \left( \sum_{i=1}^m e^{y_i/\nu} \right) = \frac{e^{\bm{y}/\nu}}{\sum_i e^{y_i/\nu}},
\end{equation}
where the division is element-wise.

Applying the chain rule, the gradient with respect to $\bm{x}$ is:
\begin{equation}
\begin{aligned}
\nabla_{\bm{x}} \hat{l}_\mathrm{stch}(\bm{x}\mid\bm{\lambda},\bm{t}) &= \nabla_{\bm{y}} \hat{l}_\mathrm{stch} \cdot \frac{\partial \bm{y}}{\partial \bm{x}} \\
&= \sum_{i=1}^m \frac{e^{y_i/\nu}}{\sum_j e^{y_j/\nu}} \nabla_{\bm{x}} \left[\lambda_i(\hat{f}_i(\bm{x}, \bm{t}) - z_i^*)\right] \\
&= \sum_{i=1}^m \frac{\lambda_i e^{y_i/\nu}}{\sum_j e^{y_j/\nu}} \nabla_{\bm{x}} \hat{f}_i(\bm{x}, \bm{t}).
\end{aligned}
\end{equation}

Define the normalized weights $w_i = \frac{\lambda_i e^{y_i/\nu}}{\sum_j e^{y_j/\nu}}$. It is straightforward to verify that $w_i \geq 0$ for all $i$ and $\sum_i w_i = \sum_i \lambda_i = 1$. Therefore:
\begin{equation}
\nabla_{\bm{x}} \hat{l}_\mathrm{stch}(\bm{x}\mid\bm{\lambda},\bm{t}) = \sum_{i=1}^m w_i \nabla_{\bm{x}} \hat{f}_i(\bm{x}; \bm{t}),
\end{equation}
which represents a weighted combination of the gradients of the LCB predictions.

For numerical stability with small smoothing parameters $\nu$, we employ the log-sum-exp trick. We first compute $\tilde{y} = \max_i y_i$ and define $\hat{y}_i = y_i - \tilde{y}$ for all $i$. This yields the numerically stable versions:
\begin{equation}
\hat{l}_\mathrm{stch}(\bm{x}\mid\bm{\lambda},\bm{t}) = \tilde{y} + \nu \log \left( \sum_{i=1}^m e^{\hat{y}_i/\nu} \right),
\end{equation}
\begin{equation}
\nabla_{\bm{x}} \hat{l}_\mathrm{stch}(\bm{x}\mid\bm{\lambda},\bm{t}) = \sum_{i=1}^m \frac{\lambda_i e^{\hat{y}_i/\nu}}{\sum_j e^{\hat{y}_j/\nu}} \nabla_{\bm{x}} \hat{f}_i(\bm{x}, \bm{t}).
\end{equation}

This formulation prevents numerical overflow while preserving the mathematical equivalence of the gradient computation.

Similarly, the gradient with respect to the hypernetwork parameters $\bm{\theta}_\mathrm{hn}$ is:
\begin{equation}\label{eq:gra_thetahn_app}
\nabla_{\bm{\theta}_\mathrm{hn}} \hat{l}_\mathrm{stch} = \frac{\partial \bm{\theta}_\mathrm{lora}}{\partial \bm{\theta}_\mathrm{hn}}\cdot \frac{\partial \bm{\theta}_\mathrm{ps}}{\partial \bm{\theta}_\mathrm{lora}} \cdot \frac{\partial h_{\bm{\theta}_\mathrm{ps}}(\bm{\lambda})}{\partial \bm{\theta}_\mathrm{ps}}\cdot \nabla_{\bm{x}} \hat{l}_\mathrm{stch}(\bm{x}\mid\bm{\lambda},\bm{t}).
\end{equation}

The Jacobian $\frac{\partial \bm{\theta}_\mathrm{lora}}{\partial \bm{\theta}_\mathrm{hn}}$ is obtained through standard backpropagation through the hypernetwork $g_{\bm{\theta}_\mathrm{hn}}$. The term $\frac{\partial \bm{\theta}_\mathrm{ps}}{\partial \bm{\theta}_\mathrm{lora}}$ follows from the LoRA parametrization in \Cref{eq:ps_theta}, where for each layer $\frac{\partial \bm{\theta}^l_\mathrm{ps}}{\partial \text{vec}(\bm{A}^l, \bm{B}^l)}$ can be obtained through appropriate Kronecker products.

All Jacobian matrices in \Cref{eq:gra_theta0_app} and \Cref{eq:gra_thetahn_app} are efficiently computed using automatic differentiation frameworks, enabling scalable gradient-based optimization of the entire PPSL-MOBO system.

\subsection{Update the Models}\label{subsec:model_update}

The learnable parameters of the base PS model are trained based on all the parameter samples, which indicates that it is used to capture the common feature of the Pareto sets with respect to different parameters. By stochastic gradient descent method, the update process can be expressed as: 
\begin{equation}\label{eq:learn_base}
    \bm{\theta}_0 \leftarrow \bm{\theta}_0 -\eta_b\nabla_{\bm{\theta}_0}\tilde{\mathcal{L}},
\end{equation}
where $\eta_b$ denotes the learning rate.
This suggests that the learnable parameters of the LoRA structure are fine-tuned on the discrepancy between the base PS model and the PS for the specific parameters. The parameters of the hypernetwork are optimized by: 
\begin{equation}\label{eq:learn_hpn}
    \bm{\theta}_{hn} \leftarrow \bm{\theta}_{hn} -\eta_{hn}\nabla_{\bm{\theta}_{hn}}\tilde{\mathcal{L}},
\end{equation}
where $\eta_{hn}$ is the step size for hypernetwork.

\section{Limitations and Broader Impact}\label{app:sec:limitation_impact}

\subsection{Limitations}

\subsubsection{Inaccuracy of the Parametric Pareto Set Model.} 
Similar to the challenges in standard PSL, the quality of PPSL-MOBO's parametric model heavily depends on the surrogate models' accuracy and the diversity of evaluated samples across the parameter space. In safety-critical applications or scenarios with limited evaluation budgets, the learned parametric mapping may not accurately capture the true Pareto-optimal solutions across all parameter configurations. This risk is amplified in parametric settings since the model must generalize not only across the decision space but also across the entire parameter space. When the parameter space is large or contains regions with drastically different objective landscapes, the hypernetwork may struggle to learn accurate mappings for underexplored parameter regions.

\subsubsection{Challenges in Preference Specification Across Parameters.} 
While PPSL-MOBO enables efficient exploration of Pareto solutions across different parameters, defining consistent preferences across varying parameter contexts remains challenging. In many real-world scenarios, user preferences may themselves be parameter-dependent, what constitutes an acceptable trade-off in one operating condition may be unsuitable in another. The current framework assumes that preferences can be specified independently of the exogenous parameters, which may not reflect the complexity of real decision-making processes. Future research could explore adaptive preference modeling that captures how trade-off preferences change with environmental parameters, potentially through interactive learning approaches that allow decision-makers to provide context-dependent feedback.

\subsubsection{Scalability in Joint Decision-Parameter Space.} 
The scalability challenge inherent in MOBO is compounded in the parametric setting, where the model must learn mappings across both high-dimensional decision spaces and parameter spaces. The hypernetwork architecture's capacity to represent complex Pareto set variations is fundamentally limited by the evaluation budget. For problems with high-dimensional parameter spaces or complex parameter-objective relationships, the current approach may require prohibitively many evaluations to achieve acceptable accuracy. Combining PPSL-MOBO with recent advances in search space decomposition methods (e.g., trust region approaches) could help focus the limited evaluation budget on the most promising regions of the joint decision-parameter space.

\subsection{Potential Societal Impact}

PPSL-MOBO's ability to efficiently learn parametric Pareto sets has significant positive implications for various domains. In engineering design, it can enable rapid exploration of design trade-offs under different operating conditions without expensive re-optimization. In personalized medicine, it could help identify treatment strategies that work across diverse patient populations. For sustainable technology development, the framework can accelerate the discovery of materials or processes that maintain performance across varying environmental conditions.

The method's real-time adaptation capability is particularly valuable for autonomous systems and adaptive control applications, where quick response to changing conditions is crucial. By pre-learning the parametric relationships, PPSL-MOBO can enable more responsive and efficient decision-making in dynamic environments, potentially improving safety and performance in applications ranging from autonomous vehicles to smart grid management.

However, the reliance on learned models also introduces risks. The parametric model's predictions should not be trusted blindly, especially in safety-critical applications where incorrect Pareto-optimal solutions could have severe consequences. The model's generalization errors in unexplored parameter regions could lead to suboptimal or even dangerous decisions if not properly validated. Additionally, the learned parametric model encodes sensitive information about the system's behavior across different conditions, which could pose security risks if exposed to adversaries in competitive or strategic settings.

To mitigate these risks, practitioners should implement appropriate validation procedures, uncertainty quantification, and fallback mechanisms when deploying PPSL-MOBO in real-world systems. The framework should be viewed as a decision support tool that augments, rather than replaces, human expertise and domain knowledge.
\section{Related Work}\label{app:sec:related_work}

\subsection{Bayesian Optimization}

Bayesian Optimization (BO) has become a cornerstone technique for tackling expensive black-box optimization problems, where objective evaluations are costly or limited in number~\cite{garnett2023bayesian}. The key idea is to construct a probabilistic surrogate model, often a Gaussian process \cite{shahriari2015taking}, that provides both mean predictions and uncertainty estimates of the true objective landscape. This surrogate is then used in conjunction with an acquisition function that strategically balances exploration and exploitation, guiding the selection of new candidate points to evaluate \cite{frazier2018tutorialbayesianoptimization, wang2023recent}.

Over the past decade, a rich body of literature has focused on advancing various facets of Bayesian optimization. Acquisition function design has been a central topic, with approaches such as Expected Improvement \cite{zhan2020expected}, Probability of Improvement, Upper Confidence Bound \cite{Srinivas2010Gaussian}, and their extensions to accommodate practical constraints \cite{Gardner2014Bayesian, Gelbart2014Bayesian} and multi-fidelity scenarios \cite{Kirthevasan2017Multi, li2020multi}. In addition, significant research has addressed the scalability of BO to high-dimensional search spaces \cite{wang2018batched}, leveraging techniques like random embeddings \cite{Wang2013Bayesian}, trust regions \cite{namura2025regional}, and additive models \cite{kandasamy2015high} to mitigate the curse of dimensionality. Batch BO algorithms have been developed to enable parallel evaluation \cite{desautels2014parallelizing}, thereby expediting the optimization process when multiple resources are available. Theoretical investigations have further provided regret bounds and convergence guarantees \cite{Kawaguchi2015Bayesian}, deepening our fundamental understanding of BO's efficiency and limitations.
For a comprehensive review of methods, applications, and theoretical foundations in Bayesian optimization, we refer the reader to Garnett~\cite{garnett2023bayesian}.

\subsection{Multi-Objective Bayesian Optimization}

Multi-Objective Bayesian Optimization (MOBO) extends the principles of single-objective Bayesian optimization to address expensive multi-objective optimization problems, where multiple conflicting objectives must be optimized simultaneously. While the Pareto set may theoretically contain an infinite number of solutions, MOBO methods typically focus on identifying a finite representative set of Pareto-optimal solutions that capture the essential trade-offs between objectives.

The landscape of MOBO algorithms can be broadly categorized into several paradigms. Scalarization-based approaches transform the multi-objective problem into a series of single-objective subproblems. ParEGO \cite{knowles2006parego} pioneered this direction by iteratively scalarizing objectives using random weight vectors and applying standard BO to each scalarized problem. TS-TCH \cite{paria2020flexible} extends this concept by incorporating Thompson sampling with Tchebycheff scalarization. MOEA/D-EGO \cite{zhang2009expensive} adopts the decomposition framework from evolutionary computation, solving multiple scalarized subproblems simultaneously with shared information among neighboring subproblems.

Another major class of MOBO algorithms directly extends acquisition functions to the multi-objective setting. SMS-EGO \cite{ponweiser2008multiobjective} and PAL \cite{zuluaga2013active, zuluaga2016pal} generalize the upper confidence bound principle to multi-objective optimization, providing theoretical guarantees on the identification of the Pareto set. The hypervolume-based acquisition functions have gained particular attention, with Emmerich et al. \cite{emmerich2006single} and Emmerich and Klinkenberg \cite{emmerich2008computation} proposing multi-objective variants of probability of improvement and expected improvement based on hypervolume contributions. These methods naturally capture the quality of a Pareto set approximation through the hypervolume indicator.

Information-theoretic approaches have also been successfully adapted to the multi-objective context. Entropy search methods, originally developed for single-objective optimization \cite{hennig2012entropy, henrandez2014predictive, wang2017max}, have been extended to multi-objective scenarios \cite{hernandez2016predictive, belakaria2019max}, offering principled ways to quantify and reduce uncertainty about the Pareto set. Bradford et al. \cite{bradford2018efficient} introduce Thompson sampling for multi-objective optimization, while Belakaria et al. \cite{belakaria2020uncertainty} propose uncertainty-based acquisition functions specifically designed for identifying diverse Pareto-optimal solutions.

Recent developments in MOBO have addressed several practical challenges and extensions. The efficient computation of hypervolume-based acquisition functions remains an active area, with Daulton et al. \cite{Daulton2020Differentiable} proposing differentiable approximations for scalable optimization. Batch MOBO methods \cite{lukovic2020diversity} enable parallel evaluations while maintaining diversity in the selected points. The handling of noisy observations \cite{daulton2022multi, tu2022joint} and high-dimensional search spaces \cite{deshwal2021combining} has received increasing attention.

A particularly relevant direction for practical applications involves incorporating decision-maker preferences into the optimization process. While some methods require preference information before or during optimization \cite{abdolshah2019multi, paria2020flexible, astudillo2019bayesian}, which may not always be feasible, recent work has begun exploring post-hoc preference elicitation and interactive approaches. However, existing MOBO methods predominantly aim to provide a finite set of approximate Pareto-optimal solutions, leaving the final selection to decision-makers without systematic support for this critical step.

\subsection{Contextual Bayesian Optimization}

Contextual Bayesian Optimization (CBO) extends traditional Bayesian Optimization by incorporating external variables, or contexts, that influence the objective function but are not directly controllable. This approach is particularly useful in scenarios where the performance of a system depends on both design parameters and environmental conditions. CBO builds upon contextual bandit frameworks \cite{Krause2011Contextual, chowdhury2017kernelized} to contextual black-box optimization via probabilistic surrogate models \cite{char2019offline, feng2020high}. Representative works in this area include offline CBO \cite{char2019offline} and continuous multi-task Bayesian optimization \cite{ginsbourger2014bayesian, pearce2018continuous}. Recent advancements in CBO have introduced frameworks like CO-BED, which employs information-theoretic principles to design contextual experiments, offering a model-agnostic solution to a variety of optimization problems \cite{ivanova2023co}. Additionally, methods such as Primal-Dual CBO have been developed to address online optimization challenges in control systems with time-average constraints, ensuring sublinear cumulative regret and zero time-average constraint violation \cite{xu2023primal}. Furthermore, approaches like Contextual Causal Bayesian Optimization integrate causal inference to identify optimal interventions, enhancing decision-making in complex systems \cite{arsenyan2023contextual}. These developments underscore the growing importance of context-aware optimization strategies in effectively navigating and optimizing complex, real-world systems.

\subsection{Pareto Set Learning}

Traditional multi-objective optimization methods have predominantly focused on finding a single solution or a finite set of Pareto-optimal solutions to approximate the Pareto set \cite{miettinen1999nonlinear, ehrgott2005multicriteria}. While these approaches have been successful in many applications, they provide only a discrete approximation of what is often a continuous Pareto set, limiting decision-makers' ability to explore the full spectrum of trade-offs between objectives.

Early attempts to approximate the entire Pareto set employed simple mathematical models \cite{rakowska1991tracing, hillermeier2001generalized, zhang2008rm, giagkiozis2014pareto}. However, these methods often struggled with complex, high-dimensional Pareto sets and were primarily developed for problems where the objective functions are known analytically.

The advent of deep learning has opened new possibilities for Pareto set approximation. \citet{pirotta2015multi} and \citet{parisi2016multi} pioneered Pareto manifold approximation in multi-objective reinforcement learning contexts. This concept has since been extended to various domains. In computer vision, preference-conditioned networks have been developed for controllable image style transfer \cite{shoshan2019dynamic, Dosovitskiy2020You}. Multi-task learning has benefited from preference-aware architectures that can either generate finite solution sets \cite{sener2018multi, lin2019pareto, mahapatra2020multi, ma2020efficient} or approximate the entire Pareto front \cite{lin2020controllable, ruchte2021scalable}. Similar advances have been made in reinforcement learning \cite{yang2019generalized, abdolmaleki2020distributional, abdolmaleki2021multi} and neural combinatorial optimization \cite{lin2022paretoiclr}.

A significant breakthrough in Pareto set learning came with the introduction of hypernetwork-based approaches \cite{chauhan2024brief}. \citet{navon2020learning} demonstrated that hypernetworks could effectively learn the mapping from preference vectors to Pareto-optimal solutions for problems with known objective functions. This approach was further refined by \cite{hoang2023improving} and \cite{tuan2024framework}, showing improved scalability and accuracy.

The extension of Pareto set learning to expensive black-box optimization represents a particularly challenging frontier. \citet{lin2022paretonips} introduced PSL-MOBO, the first method to apply Pareto set learning principles to black-box multi-objective optimization. This approach leverages surrogate models, typically Gaussian processes, to learn the preference-to-solution mapping without direct access to the true objective functions. By learning a continuous representation of the Pareto set, PSL-MOBO enables decision-makers to explore any trade-off preference after the optimization process, without requiring additional expensive function evaluations.

\subsection{Learning to Optimize}

Learning to optimize represents a paradigm shift from traditional optimization methods that treat each problem in isolation. Instead of hand-crafting optimization algorithms, this approach trains models, typically neural networks, on large collections of optimization tasks to learn effective optimization strategies that can generalize to new problems. By internalizing patterns and structures from previously solved tasks, these learned optimizers can rapidly adapt to new optimization challenges without starting from scratch.

Recent advances have explored various ways to incorporate learning into the optimization process. Some approaches focus on learning components of traditional optimizers, such as amortizing the surrogate model \cite{muller2023pfns4bo, chang2025amortized} or the acquisition function \cite{swersky2020amortized, Volpp2020Meta} in Bayesian optimization. These methods retain the overall structure of classical BO while enhancing specific components through learned representations. More ambitious approaches pursue fully end-to-end learning systems \cite{chen2017learning, chen2022towards, yang2024mongoose, maraval2023end} that directly learn to map from problem observations to optimization decisions, bypassing the need for explicit surrogate models and acquisition functions entirely.

The success of learned optimizers depends heavily on architectural choices that capture the inherent structure of optimization problems. Conditional neural processes \cite{Garnelo2018Conditional}, built upon the Deep Sets framework \cite{zaheer2017deep}, have emerged as particularly suitable for optimization tasks \cite{maraval2023end, muller2023pfns4bo}. Their permutation-invariant design naturally handles the unordered nature of function evaluations collected during optimization. Transformer-based neural processes \cite{nguyen2022transformer, muller2022transformers, chang2025amortized} further enhance this capability through attention mechanisms that model complex dependencies between observations, enabling richer representations of the optimization landscape.

Despite significant progress in learning to optimize for single-objective problems, the extension to multi-objective settings remains largely unexplored. The added complexity of handling multiple conflicting objectives, learning appropriate trade-off strategies, and incorporating preference information poses substantial challenges that existing methods have yet to address. Our work tackles this gap by developing a learning-based approach specifically designed for multi-objective optimization, where the model learns directly from preference-conditioned data to efficiently navigate the Pareto frontier of new problems while providing comprehensive coverage of all possible trade-offs.

\section{Experimental Analysis: Multi-Objective Optimization with Shared Components}\label{app:sec:experiment_mopsc}

\subsection{Experimental Setups}

\subsubsection{Optimization Problems.}
To evaluate the performance of our method, we use the real-world multiobjective test problems from the RE suite \cite{tanabe2020easy}. This test suite is well-suited for our study as it contains problems with diverse characteristics and Pareto front geometries. The implementation details and problem definitions can be found in the repository.

\begin{table}[h]
\small
    \centering
    \begin{tabular}{l|c}
    \toprule[1.0pt]    
        Hyperparameter & Value \\
        \midrule
        \# layers for hypernetwork & 4 \\
        \# neurons for hypernetwork & 1024 \\ 
        \# layers for PS model & 2 for $m=2$ and 3 for $m=3$ \\
        \# neurons for PS model & 512 for $m=2$ and 256 for $m=3$ \\
        Rank $r$ & 3 \\
        Learning rate $\eta_{\mathrm{hn}}$ & $10^{-5}$ \\
        Learning rate $\eta_{\mathrm{b}}$ & $10^{-3}$ \\
        \# samples for parameter $N_t$ & 20 \\ 
        \# samples for weight vector $N_{\lambda}$ & 10 \\
        Smoothing parameter $\nu$ in STCH & 0.01 \\
        Coefficient $\beta$ in LCB function & 0.05 \\
        Batch size $B$ in acquisition & 5 \\ 
        \bottomrule[1.0pt]
    \end{tabular}
    \caption{Hyperparameters setting for PPSL-MOBO.}
    \label{tab:hyperpara}
\end{table}

\subsubsection{Baseline Methods.}
To evaluate the efficiency and efficacy of our proposed PPSL-MOBO method, we conduct comprehensive comparisons against three categories of baseline algorithms: (1) Multi-objective evolutionary algorithm: NSGA-II \cite{deb2002fast}, a widely adopted MOEA that serves as a representative evolutionary approach; (2) Standard MOBO methods: qParEGO \cite{knowles2006parego} and qEHVI \cite{Daulton2020Differentiable}, two well-established multi-objective Bayesian optimization algorithms recognized for their sample efficiency, which employ Gaussian Process models and utilize Expected Improvement and Hypervolume Improvement acquisition functions, respectively; and (3) Model-based MOBO: PSL-MOBO \cite{lin2022paretonips}, a recent approach that learns a non-parametric model of the Pareto set within the Bayesian optimization framework. All experiments are implemented using the pymoo library for evolutionary algorithms, BoTorch for MOBO methods, and the PSL-MOBO code repository.

\subsubsection{Hyperparameter setting.}
We present the hyperparameter configuration for PPSL-MOBO in Table \ref{tab:hyperpara}. The architecture employs a 4-layer hypernetwork with 1024 neurons per layer to generate the low-rank adaptation parameters. The PS model uses a shallower architecture with 2 hidden layers (512 neurons each) for bi-objective problems and 3 hidden layers (256 neurons each) for tri-objective problems, reflecting the increased complexity of higher-dimensional Pareto sets. We set the LoRA rank $r=3$, which provides sufficient expressiveness while maintaining computational efficiency. The learning rates are set differently for the hypernetwork ($\eta_\mathrm{hn} = 10^{-5}$) and base parameters ($\eta_\mathrm{b} = 10^{-3}$) to account for their different convergence characteristics. During training, we sample $N_t=20$ task parameters and $N_{\lambda}=10$ preference vectors per parameter to approximate the expectation in the training objective. The STCH smoothing parameter is set to 0.01 for numerical stability, while the LCB coefficient $\beta=0.05$ balances exploration and exploitation in the surrogate models. For data acquisition, we select batches of $B=5$ points per iteration to parallelize expensive function evaluations.

\subsubsection{Performance Indicator.}
The hypervolume (HV) \cite{zitzler1999multiobjective} indicator is used to measure the quality of the obtained Pareto front approximations. To ensure a fair comparison, all solutions are first normalized into the unit hypercube $^m$ using the pre-computed ideal and nadir points provided for the RE test suite \cite{tanabe2020easy}. The HV is then calculated with a reference point of $(1.1, \ldots, 1.1)^\top$.

Since no existing method is specifically designed for this parametric MOP formulation, we compare against strong baseline optimizers that are re-executed from scratch for each given shared component value. We randomly sample ten distinct shared component values for each problem. The final HV scores are reported as the average over these ten tasks and three independent runs per task.

For a fair comparison of sample efficiency, each baseline method (NSGA-II, qParEGO, qEHVI, and PSL-MOBO) is allocated a budget of 100 expensive function evaluations for each set of shared component values.
In contrast, our PPSL-MOBO method is trained once with a total budget of 200 expensive function evaluations. This budget is used to build the surrogate models and train the parametric PS model across the entire space of shared components. After this one-time training phase, PPSL-MOBO can generate the approximate Pareto set for any new shared component value via inference, which incurs negligible computational cost.

\subsection{Experimental Results}\label{app:mopsc_res}

\begin{table*}
\scriptsize
\centering
\begin{tabular}{lc|ccccc}
    \toprule[1.0pt]
    Problem & Shared Comp. & NSGA-II & qParEGO & qEHVI & PSL-MOBO & PPSL-MOBO \\
    \midrule\midrule
RE33 & $(x_1)$ & 6.68e-01 (6.38e-03) & 7.93e-01 (7.16e-03) & 8.38e-01 (3.13e-04) & \cellcolor{gray!30}\textbf{8.40e-01} (5.15e-03) & 6.99e-01 (2.14e-01) \\
 & $(x_2)$ & 6.64e-01 (4.53e-03) & 7.65e-01 (4.27e-03) & \cellcolor{gray!30}\textbf{8.15e-01} (1.23e-03) & 7.88e-01 (4.46e-03) & 8.07e-01 (1.21e-02) \\
 & $(x_3)$ & 4.00e-01 (6.91e-03) & 4.94e-01 (3.65e-03) & 5.14e-01 (4.14e-03) & 4.98e-01 (2.77e-03) & \cellcolor{gray!30}\textbf{5.28e-01} (7.28e-03) \\
 & $(x_4)$ & 6.75e-01 (8.93e-03) & 8.06e-01 (3.42e-03) & 8.20e-01 (2.43e-03) & 7.77e-01 (1.55e-02) & \cellcolor{gray!30}\textbf{8.36e-01} (8.61e-03) \\
 & $(x_1,x_2)$ & 4.50e-01 (4.13e-03) & 4.80e-01 (5.20e-03) & \cellcolor{gray!30}\textbf{4.91e-01} (2.31e-05) & 4.83e-01 (1.32e-03) & 4.86e-01 (5.51e-05) \\
 & $(x_2,x_3)$ & 4.50e-01 (2.55e-03) & 4.80e-01 (1.83e-03) & \cellcolor{gray!30}\textbf{4.93e-01} (2.09e-05) & 4.78e-01 (4.02e-04) & \cellcolor{gray!30}\textbf{4.93e-01} (2.11e-03) \\
 & $(x_3,x_4)$ & 3.54e-01 (1.97e-03) & 3.80e-01 (2.26e-03) & 3.85e-01 (1.08e-03) & 3.82e-01 (3.44e-04) & \cellcolor{gray!30}\textbf{3.93e-01} (1.20e-04) \\
 & $(x_1,x_2,x_3)$ & 2.17e-01 (9.37e-04) & 2.20e-01 (6.08e-04) & \cellcolor{gray!30}\textbf{2.21e-01} (5.03e-06) & 2.15e-01 (5.81e-04) & \cellcolor{gray!30}\textbf{2.21e-01} (8.16e-05) \\
 & $(x_2,x_3,x_4)$ & 8.61e-02 (9.38e-04) & 8.85e-02 (1.45e-04) & \cellcolor{gray!30}\textbf{8.86e-02} (1.85e-05) & 8.08e-02 (6.83e-03) & 7.26e-02 (1.48e-02) \\
     \midrule
 RE37 & $(x_1)$ & 4.43e-01 (8.12e-03) & 5.25e-01 (3.21e-03) & \cellcolor{gray!30}\textbf{5.70e-01} (4.44e-04) & 5.57e-01 (1.19e-03) & 5.28e-01 (3.66e-02) \\
 & $(x_2)$ & 4.00e-01 (4.04e-03) & 4.68e-01 (3.83e-03) & 5.02e-01 (9.17e-04) & 4.99e-01 (1.58e-03) & \cellcolor{gray!30}\textbf{5.08e-01} (9.64e-05) \\
 & $(x_3)$ & 4.67e-01 (5.14e-03) & 6.02e-01 (2.83e-03) & \cellcolor{gray!30}\textbf{6.36e-01} (4.38e-05) & 6.14e-01 (7.48e-03) & \cellcolor{gray!30}\textbf{6.36e-01} (6.03e-03) \\
 & $(x_4)$ & 4.12e-01 (7.74e-03) & 4.98e-01 (4.80e-03) & \cellcolor{gray!30}\textbf{5.46e-01} (4.54e-04) & 4.97e-01 (1.43e-02) & 4.92e-01 (1.40e-03) \\
 & $(x_1,x_2)$ & 2.92e-01 (2.94e-03) & 3.07e-01 (2.08e-03) & 3.26e-01 (1.87e-05) & 3.25e-01 (1.99e-04) & \cellcolor{gray!30}\textbf{3.27e-01} (1.27e-04) \\
 & $(x_2,x_3)$ & 3.33e-01 (1.98e-03) & 3.79e-01 (3.09e-03) & \cellcolor{gray!30}\textbf{3.85e-01} (1.48e-05) & 3.81e-01 (6.94e-04) & \cellcolor{gray!30}\textbf{3.85e-01} (2.58e-05) \\
 & $(x_3,x_4)$ & 3.57e-01 (5.59e-03) & 4.02e-01 (1.10e-04) & \cellcolor{gray!30}\textbf{4.12e-01} (5.53e-06) & 3.91e-01 (4.86e-03) & 4.04e-01 (1.26e-02) \\
 & $(x_1,x_2,x_3)$ & 2.10e-01 (8.64e-04) & \cellcolor{gray!30}\textbf{2.15e-01} (2.76e-05) & \cellcolor{gray!30}\textbf{2.15e-01} (4.85e-06) & \cellcolor{gray!30}\textbf{2.15e-01} (3.93e-06) & \cellcolor{gray!30}\textbf{2.15e-01} (1.71e-05) \\
 & $(x_2,x_3,x_4)$ & 2.21e-01 (5.20e-04) & \cellcolor{gray!30}\textbf{2.25e-01} (4.43e-05) & \cellcolor{gray!30}\textbf{2.25e-01} (1.10e-05) & 2.23e-01 (1.30e-04) & \cellcolor{gray!30}\textbf{2.25e-01} (3.18e-05) \\
     \midrule
 RE61 & $(x_1)$ & 5.36e-01 (5.93e-03) & 6.23e-01 (6.93e-04) & 6.38e-01 (6.47e-05) & 6.39e-01 (6.51e-04) & \cellcolor{gray!30}\textbf{6.42e-01} (2.12e-03) \\
 & $(x_2)$ & 5.26e-01 (1.17e-03) & 5.65e-01 (7.49e-04) & 5.68e-01 (7.94e-04) & 5.66e-01 (6.01e-04) & \cellcolor{gray!30}\textbf{5.70e-01} (7.34e-04) \\
 & $(x_3)$ & 5.20e-01 (7.08e-03) & 6.70e-01 (1.44e-03) & 6.79e-01 (4.27e-03) & 6.93e-01 (4.89e-04) & \cellcolor{gray!30}\textbf{7.06e-01} (7.43e-04) \\
    \bottomrule[1.0pt]
\end{tabular}
\caption{HV values of PPSL-MOBO and baseline approaches on MOPs with shared components.All baseline methods were re-executed with a budget of 100 evaluations per task across the ten tested problems. In contrast, PPSL-MOBO was trained only once, using a total of 200 evaluations, and subsequently inferred the PS for each parameterized task.}
\label{tab:mopsc_compare_baseline_addi}
\end{table*}

Table \ref{tab:mopsc_compare_baseline_addi} presents the hypervolume (HV) values achieved by PPSL-MOBO and baseline methods on three additional real-world engineering problems (RE33, RE37, and RE61) with various shared component configurations. The results reveal several key insights about the performance and efficiency of our proposed approach.

\paragraph{Overall Performance.} PPSL-MOBO demonstrates highly competitive performance across different shared component scenarios, achieving the best HV values in 11 out of 18 test cases (highlighted in gray). Notably, PPSL-MOBO shows particularly strong performance when multiple components are shared simultaneously, such as in RE33 with $(x_3, x_4)$ and RE37 with $(x_1, x_2)$ and $(x_2, x_3)$ configurations. This suggests that our hypernetwork-LoRA architecture effectively captures the complex interactions between shared components and their impact on the Pareto set structure.

\paragraph{Sample Efficiency.} A critical advantage of PPSL-MOBO is its remarkable sample efficiency in handling parametric optimization problems. In our experimental setup, for each of the nine shared component configurations, we randomly sampled ten parameter values to evaluate the methods' ability to generalize across the parameter space. All the other baseline methods must re-execute the entire optimization process for each parameter value, requiring 100 evaluations per parameter. This results in 1,000 evaluations for ten parameter values per configuration, and the computational cost scales linearly with the number of parameters to be tested. In contrast, PPSL-MOBO uses only 200 evaluations in total to train a unified parametric model that can instantly generate the Pareto set for any parameter value within the shared component configuration. This represents a 50× reduction in function evaluations when considering ten parameter values, and the advantage becomes even more pronounced as the number of query parameters increases. This efficiency is particularly valuable in real-world applications where each evaluation may involve costly simulations, physical experiments, or time-sensitive decisions requiring real-time adaptation.

\begin{figure}[ht]
    \centering
    \includegraphics[width=0.99\linewidth]{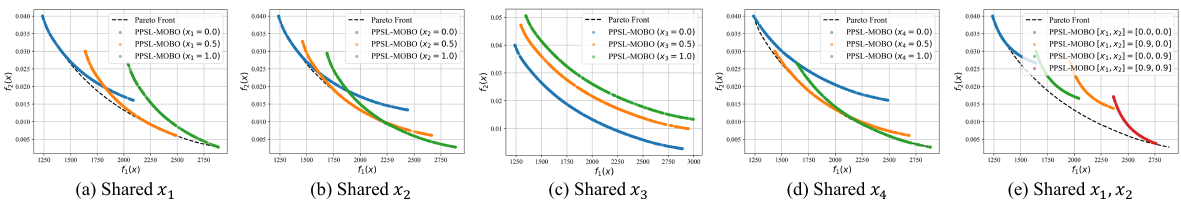}
    \caption{Learned Pareto fronts for the Four-Bar Truss Design Problem (RE21) with shared components. The dashed black line represents the unconstrained Pareto front. The color coding indicates different parameter values, illustrating how the Pareto front geometry evolves as the shared component values change.}
    \label{fig:ps_re21}
\end{figure}

\begin{figure}[ht]
    \centering
    \includegraphics[width=0.99\linewidth]{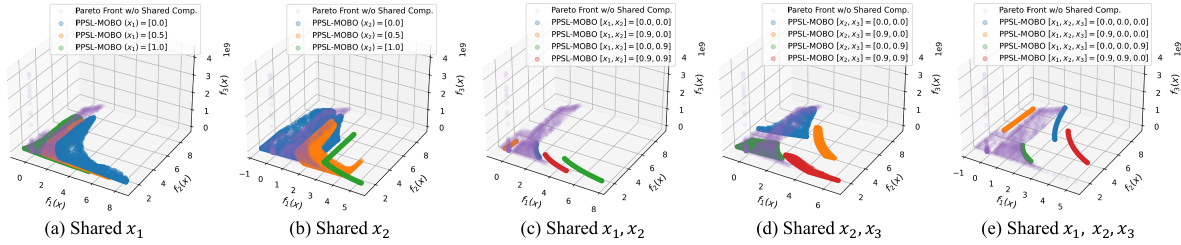}
    \caption{Learned Pareto fronts for the Disc Brake Design Problem (RE33) with shared components.}
    \label{fig:ps_re33}
\end{figure}

\paragraph{Computational Time Analysis.} Table \ref{tab:mopsc_runtime_compare} provides a detailed comparison of computational times between PPSL-MOBO and baseline model-based methods. The results reveal the substantial time savings achieved by our approach. While baseline methods require 317-2,669 seconds for optimizing a shared component configuration, PPSL-MOBO's training time ranges from 364-639 seconds, which is comparable to or faster than most baseline methods despite learning a more complex parametric model. The most striking advantage emerges during inference: PPSL-MOBO can generate a Pareto set with 5,000 solutions for a new parameter value in just 2-4 milliseconds, representing a speedup of approximately 10,000-100,000× compared to re-running baseline methods. This millisecond-level inference capability is crucial for applications requiring real-time decision-making or rapid design space exploration. Furthermore, while baseline methods' total time scales linearly with the number of parameter queries (e.g., 10× longer for 10 parameters), PPSL-MOBO's inference time remains constant, making it increasingly advantageous as the number of parameter variations grows.

\begin{table*}[h]
\scriptsize
\centering
\begin{tabular}{l|ccccc}
\toprule[1.0pt]
Problem & qParEGO & qEHVI & PSL-MOBO & PPSL-MOBO Training & PPSL-MOBO Inference \\
\midrule\midrule
RE21 & 317.1s & 551.3s & 1761.3s & 364.4s & 0.019s \\
RE33 & 430.9s & 1038.9s & 1812.4s & 581.8s & 0.040s \\
RE37 & 349.2s & 873.1s & 2668.9s & 639.0s & 0.028s \\
\bottomrule[1.0pt]
\end{tabular}
\caption{Run time comparison (in seconds) of PPSL-MOBO and baseline model-based methods. All methods are run on the same CPU.}
\label{tab:mopsc_runtime_compare}
\end{table*}

\paragraph{Robustness Analysis.} The standard deviations (shown in parentheses of \Cref{tab:mopsc_compare_baseline_addi}) indicate that PPSL-MOBO generally maintains stable performance across different runs. In some cases, such as RE33 with $(x_1)$, PPSL-MOBO shows higher variance, which can be attributed to the stochastic nature of the GP-based acquisition process and the challenge of learning a parametric model that must generalize across diverse shared component configurations.

\paragraph{Comparison with Baselines.} Among the baseline methods, qEHVI typically achieves the strongest performance, followed by PSL-MOBO and qParEGO, while NSGA-II consistently shows the lowest HV values. This ranking aligns with expectations, as MOBO methods are designed for sample-efficient optimization. However, even the best-performing baseline methods require re-optimization for each new shared component configuration, whereas PPSL-MOBO provides instant inference after the initial training phase.

\begin{table}[]
    \scriptsize
    \centering
    \begin{tabular}{lc|ccccc}
    \toprule
        Problem & Shared Comp. & $\beta=0.01$ & $\beta=0.05$* & $\beta=0.1$ & $\beta=0.5$ & $\beta=1$ \\
        \midrule\midrule
RE21 & $(x_1)$ & 7.340e-01 (2.183e-05) & \cellcolor{gray!30}\textbf{7.341e-01} (3.975e-05) & 7.340e-01 (1.564e-05) & \cellcolor{gray!30}\textbf{7.340e-01} (9.659e-05) & 7.341e-01 (2.040e-05) \\
 & $(x_2)$ & \cellcolor{gray!30}\textbf{7.852e-01} (8.294e-05) & \cellcolor{gray!30}\textbf{7.852e-01} (3.589e-05) & \cellcolor{gray!30}\textbf{7.852e-01} (3.940e-05) & \cellcolor{gray!30}\textbf{7.852e-01} (2.676e-05) & \cellcolor{gray!30}\textbf{7.852e-01} (5.565e-05) \\
 & $(x_3)$ & \cellcolor{gray!30}\textbf{6.771e-01} (6.221e-05) & \cellcolor{gray!30}\textbf{6.771e-01} (1.510e-05) & \cellcolor{gray!30}\textbf{6.771e-01} (7.312e-05) & 6.770e-01 (2.374e-05) & \cellcolor{gray!30}\textbf{6.771e-01} (7.208e-05) \\
 & $(x_4)$ & \cellcolor{gray!30}\textbf{8.025e-01} (3.444e-05) & \cellcolor{gray!30}\textbf{8.025e-01} (3.812e-05) & 8.024e-01 (1.122e-04) & \cellcolor{gray!30}\textbf{8.025e-01} (3.111e-05) & \cellcolor{gray!30}\textbf{8.025e-01} (1.066e-04) \\
 & $(x_1,x_2)$ & \cellcolor{gray!30}\textbf{6.235e-01} (1.710e-05) & \cellcolor{gray!30}\textbf{6.235e-01} (3.752e-06) & \cellcolor{gray!30}\textbf{6.235e-01} (1.309e-05) & \cellcolor{gray!30}\textbf{6.235e-01} (1.235e-05) & \cellcolor{gray!30}\textbf{6.235e-01} (5.614e-06) \\
 & $(x_2,x_3)$ & 5.681e-01 (5.969e-05) & \cellcolor{gray!30}\textbf{5.682e-01} (2.079e-05) & 5.681e-01 (7.843e-05) & \cellcolor{gray!30}\textbf{5.682e-01} (2.320e-05) & \cellcolor{gray!30}\textbf{5.682e-01} (8.711e-06) \\
 & $(x_3,x_4)$ & 5.656e-01 (4.726e-05) & 5.657e-01 (5.439e-05) & 5.657e-01 (5.988e-05) & 5.657e-01 (5.488e-05) & \cellcolor{gray!30}\textbf{5.658e-01} (5.255e-05) \\
 & $(x_1,x_2,x_3)$ & \cellcolor{gray!30}\textbf{4.000e-01} (1.844e-05) & \cellcolor{gray!30}\textbf{4.000e-01} (2.947e-05) & \cellcolor{gray!30}\textbf{4.000e-01} (2.482e-05) & \cellcolor{gray!30}\textbf{4.000e-01} (1.219e-06) & \cellcolor{gray!30}\textbf{4.000e-01} (1.576e-05) \\
 & $(x_2,x_3,x_4)$ & \cellcolor{gray!30}\textbf{5.184e-01} (2.468e-05) & \cellcolor{gray!30}\textbf{5.184e-01} (4.868e-05) & \cellcolor{gray!30}\textbf{5.184e-01} (2.062e-05) & \cellcolor{gray!30}\textbf{5.184e-01} (1.655e-05) & \cellcolor{gray!30}\textbf{5.184e-01} (1.665e-05) \\
 \midrule
RE33 & $(x_1)$ & \cellcolor{gray!30}\textbf{8.478e-01} (1.250e-03) & 7.999e-01 (6.280e-02) & 8.417e-01 (8.543e-03) & 8.449e-01 (3.239e-03) & 8.448e-01 (4.530e-03) \\
 & $(x_2)$ & 8.188e-01 (5.363e-03) & 8.186e-01 (4.517e-03) & 7.168e-01 (1.482e-01) & \cellcolor{gray!30}\textbf{8.191e-01} (3.282e-03) & 8.108e-01 (9.111e-03) \\
 & $(x_3)$ & \cellcolor{gray!30}\textbf{5.309e-01} (4.842e-04) & 5.252e-01 (8.064e-03) & 5.308e-01 (4.636e-04) & 5.302e-01 (1.436e-03) & 5.081e-01 (2.117e-02) \\
 & $(x_4)$ & 8.348e-01 (4.237e-03) & 8.326e-01 (9.191e-03) & 8.348e-01 (9.694e-03) & \cellcolor{gray!30}\textbf{8.434e-01} (1.547e-03) & 8.203e-01 (3.255e-02) \\
 & $(x_1,x_2)$ & \cellcolor{gray!30}\textbf{4.898e-01} (1.843e-03) & 4.849e-01 (6.920e-03) & 3.981e-01 (1.314e-01) & 4.862e-01 (6.844e-04) & 4.854e-01 (2.154e-04) \\
 & $(x_2,x_3)$ & 4.745e-01 (2.666e-02) & 4.555e-01 (5.350e-02) & \cellcolor{gray!30}\textbf{4.934e-01} (3.899e-04) & 4.931e-01 (2.090e-04) & 4.930e-01 (5.387e-04) \\
 & $(x_3,x_4)$ & \cellcolor{gray!30}\textbf{3.898e-01} (3.407e-03) & 3.564e-01 (3.966e-02) & 3.896e-01 (2.402e-03) & 3.864e-01 (1.227e-03) & 3.892e-01 (2.177e-03) \\
 & $(x_1,x_2,x_3)$ & \cellcolor{gray!30}\textbf{2.212e-01} (1.614e-05) & \cellcolor{gray!30}\textbf{2.212e-01} (2.208e-05) & 2.211e-01 (4.869e-05) & 2.211e-01 (7.756e-05) & \cellcolor{gray!30}\textbf{2.212e-01} (6.121e-05) \\
 & $(x_2,x_3,x_4)$ & 8.788e-02 (7.541e-04) & 8.880e-02 (2.678e-04) & 8.763e-02 (1.860e-03) & 7.668e-02 (1.693e-02) & \cellcolor{gray!30}\textbf{8.894e-02} (2.044e-05) \\
    \bottomrule
    \end{tabular}
    \caption{Comparison of PPSL-MOBO between different LCB coefficient $\beta$. $\beta$ controls the balance between exploration and exploitation. Lower $\beta$ value emphasizes on explitation.}
    \label{tab:hyper_sensitivity_beta}
\end{table}

\begin{table}[h]
    \scriptsize
    \centering
    \begin{tabular}{lc|cccc}
    \toprule
        Problem & Shared Comp. & $\nu=0.001$ & $\nu=0.01$* & $\nu=0.1$ & $\nu=1.0$ \\
        \midrule\midrule
RE21 & $(x_1)$ & \cellcolor{gray!30}\textbf{7.340e-01} (3.773e-05) & \cellcolor{gray!30}\textbf{7.341e-01} (1.308e-05) & \cellcolor{gray!30}\textbf{7.341e-01} (4.722e-05) & \cellcolor{gray!30}\textbf{7.341e-01} (2.864e-05) \\
 & $(x_2)$ & 7.853e-01 (1.585e-05) & 7.852e-01 (1.899e-05) & 7.853e-01 (1.384e-05) & \cellcolor{gray!30}\textbf{7.854e-01} (4.434e-05) \\
 & $(x_3)$ & 6.771e-01 (7.254e-05) & \cellcolor{gray!30}\textbf{6.772e-01} (6.070e-05) & \cellcolor{gray!30}\textbf{6.772e-01} (1.353e-04) & \cellcolor{gray!30}\textbf{6.772e-01} (1.242e-05) \\
 & $(x_4)$ & \cellcolor{gray!30}\textbf{8.025e-01} (3.036e-06) & \cellcolor{gray!30}\textbf{8.025e-01} (2.991e-05) & \cellcolor{gray!30}\textbf{8.025e-01} (6.521e-06) & 8.024e-01 (1.156e-04) \\
 & $(x_1,x_2)$ & 6.235e-01 (6.551e-06) & 6.235e-01 (1.140e-05) & 6.235e-01 (1.863e-05) & \cellcolor{gray!30}\textbf{6.236e-01} (1.022e-05) \\
 & $(x_2,x_3)$ & 5.682e-01 (6.511e-05) & 5.683e-01 (3.704e-05) & 5.683e-01 (6.929e-05) & \cellcolor{gray!30}\textbf{5.685e-01} (3.829e-05) \\
 & $(x_3,x_4)$ & 5.657e-01 (6.269e-05) & \cellcolor{gray!30}\textbf{5.658e-01} (3.136e-05) & \cellcolor{gray!30}\textbf{5.658e-01} (1.404e-05) & \cellcolor{gray!30}\textbf{5.658e-01} (3.865e-05) \\
 & $(x_1,x_2,x_3)$ & 4.000e-01 (3.580e-05) & 4.000e-01 (1.251e-05) & 4.000e-01 (6.750e-06) & \cellcolor{gray!30}\textbf{4.002e-01} (5.463e-06) \\
 & $(x_2,x_3,x_4)$ & 5.184e-01 (3.369e-05) & 5.184e-01 (1.310e-05) & 5.184e-01 (6.497e-05) & \cellcolor{gray!30}\textbf{5.185e-01} (1.318e-05) \\
         \midrule
RE33 & $(x_1)$ & 8.463e-01 (2.591e-04) & 8.470e-01 (1.765e-03) & \cellcolor{gray!30}\textbf{8.476e-01} (1.136e-03) & 8.446e-01 (4.149e-03) \\
 & $(x_2)$ & 8.134e-01 (5.090e-03) & 7.822e-01 (1.214e-02) & 8.169e-01 (4.877e-03) & \cellcolor{gray!30}\textbf{8.204e-01} (9.541e-04) \\
 & $(x_3)$ & 5.275e-01 (3.624e-03) & \cellcolor{gray!30}\textbf{5.301e-01} (4.506e-04) & 5.288e-01 (2.515e-03) & 5.160e-01 (3.237e-03) \\
 & $(x_4)$ & 8.336e-01 (9.267e-03) & 8.318e-01 (9.737e-03) & \cellcolor{gray!30}\textbf{8.355e-01} (1.036e-02) & 8.293e-01 (1.476e-02) \\
 & $(x_1,x_2)$ & \cellcolor{gray!30}\textbf{4.896e-01} (2.995e-03) & 4.894e-01 (2.765e-03) & 4.878e-01 (3.232e-03) & 4.848e-01 (8.223e-04) \\
 & $(x_2,x_3)$ & \cellcolor{gray!30}\textbf{4.933e-01} (6.053e-04) & 4.904e-01 (1.786e-04) & 4.927e-01 (1.015e-03) & 4.149e-01 (4.953e-02) \\
 & $(x_3,x_4)$ & 3.856e-01 (3.447e-03) & 3.899e-01 (3.546e-03) & \cellcolor{gray!30}\textbf{3.913e-01} (8.256e-04) & 3.865e-01 (1.297e-03) \\
 & $(x_1,x_2,x_3)$ & 2.211e-01 (5.413e-05) & \cellcolor{gray!30}\textbf{2.212e-01} (2.343e-05) & \cellcolor{gray!30}\textbf{2.212e-01} (3.909e-05) & \cellcolor{gray!30}\textbf{2.212e-01} (3.437e-05) \\
 & $(x_2,x_3,x_4)$ & 7.681e-02 (1.701e-02) & \cellcolor{gray!30}\textbf{7.687e-02} (1.706e-02) & 7.578e-02 (1.635e-02) & 7.599e-02 (1.644e-02) \\
    \bottomrule
    \end{tabular}
    \caption{Comparison of PPSL-MOBO between different STCH smooth coefficient $\nu$. Larger $\nu$ represents more accurate TCH approximation.}
    \label{tab:hyper_sensitivity_nu}
\end{table}

\paragraph{Parametric Pareto Front Visualization.} To demonstrate the effectiveness of our learned parametric model, we select specific shared component configurations and choose several specific parameter values within each configuration to visualize the resulting Pareto fronts, as shown in \Cref{fig:ps_re21} and \Cref{fig:ps_re33}. These figures provide visual insights into how PPSL-MOBO captures the parametric evolution of Pareto fronts under different shared component constraints. For the bi-objective RE21 problem, we present 2D Pareto fronts, while for the tri-objective RE33 problem, we show 3D visualizations. We observe that constraining different components leads to distinct transformations of the Pareto front geometry. For instance, in RE21 with shared $x_1$, the Pareto fronts maintain similar shapes but shift positions as the parameter value changes from 0.0 to 1.0. In RE33, the 3D visualizations reveal the complex manifold structures formed by these parametric Pareto fronts, where each colored surface represents the Pareto front for a specific parameter value. Notably, when multiple components are shared (subfigures (c)-(e) in RE21 and (c)-(e) in RE33), the Pareto fronts exhibit more constrained geometries with reduced spread, reflecting the decreased degrees of freedom in the optimization problem. The smooth transitions between Pareto fronts for different parameter values demonstrate that PPSL-MOBO successfully learns the continuous mapping from shared component values to their corresponding Pareto sets, enabling decision-makers to understand how design constraints affect the trade-off landscape without requiring repeated optimizations.

\begin{figure}[h]
    \centering
    \includegraphics[width=0.98\linewidth]{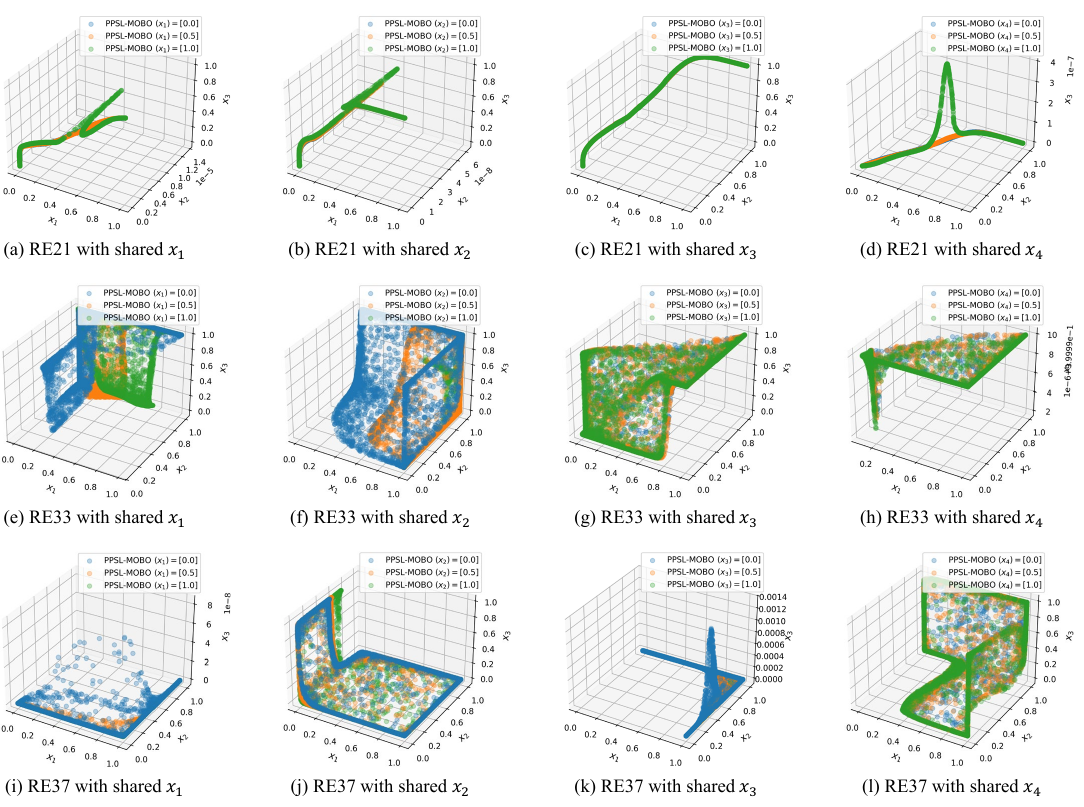}
    \caption{Learned Pareto sets for RE21, RE33, and Re37 problems with one shared components.}
    \label{fig:mopsc_ps}
\end{figure}

\paragraph{Parametric Pareto Set Visualization.} To demonstrate the effectiveness of our learned parametric model, we visualize the learned Pareto sets under varying contextual parameters in \Cref{fig:mopsc_ps}. The visualizations illustrate that the model successfully captures diverse underlying geometries of the Pareto sets, which range from simple, low-dimensional manifolds to complex, high-dimensional surfaces. Crucially, the plots, which color-code the sets according to different parameter values, reveal a smooth and systematic transformation of the Pareto set in the objective space as the shared component is varied. This confirms that our parametric approach effectively learns a continuous mapping from the contextual parameters to the entire family of corresponding Pareto sets, enabling efficient optimization across diverse problem instances.

\subsection{Hyperparameter Sensitivity}

To evaluate the robustness of our framework and provide guidance for its application, we conduct a sensitivity analysis on several key hyperparameters. This section examines the impact of the LCB coefficient $\beta$, the smoothing parameter $\nu$ in the STCH scalarization function, and the rank $r$ of the LoRA matrices on the model's performance.

First, we analyze the LCB coefficient $\beta$, which governs the trade-off between exploitation (using the GP mean) and exploration (using the GP uncertainty) during the surrogate-based training phase. \cref{tab:hyper_sensitivity_beta} presents the hypervolume (HV) performance of PPSL-MOBO on the RE21 and RE33 problems across a range of $\beta$ values: $\{0.01, 0.05, 0.1, 0.5, 1\}$. The results indicate that the algorithm's performance is remarkably robust on the RE21 problem, with HV scores remaining stable and showing negligible variation across all tested $\beta$ values. This suggests that for certain problem landscapes, the final learned Pareto set model is not highly sensitive to the exploration-exploitation balance. Conversely, the performance on the RE33 problem demonstrates greater sensitivity, where the optimal $\beta$ value varies depending on the specific shared component configuration. For instance, a more exploitative strategy ($\beta=0.01$) yields the best result for the ($x_1$) configuration, whereas a more explorative one ($\beta=0.5$) is optimal for ($x_2$). This variability highlights that the ideal balance is problem-dependent. Nevertheless, the chosen default value of $\beta=0.05$ (marked with an asterisk) proves to be a well-balanced and reliable choice, as it consistently delivers competitive performance across most test cases without being an extreme selection.

We then examine the impact of the STCH smoothing coefficient $\nu$, which controls the trade-off between the smoothness of the scalarization function and its accuracy in approximating the non-smooth Tchebycheff function. \Cref{tab:hyper_sensitivity_nu} presents the results for $\nu$ values of $\{0.001, 0.01, 0.1, 1.0\}$. The findings mirror the previous analysis, showing that the framework's performance is largely robust to this hyperparameter. On the RE21 problem, the HV scores remain remarkably stable across all tested values of $\nu$. For the RE33 problem, there is greater sensitivity, with the optimal choice of $\nu$ varying with the shared component configuration. Overall, our selected default value of $\nu=0.01$ (marked with an asterisk) proves to be a well-justified and balanced choice, delivering consistently strong performance across the majority of test cases.

\begin{table}[h]
    \scriptsize
    \centering
    \begin{tabular}{lc|ccc}
    \toprule
        Problem & Shared Comp. & $r=2$ & $r=3$* & $r=5$ \\
        \midrule\midrule
RE21 & $(x_1)$ & \cellcolor{gray!30}\textbf{7.340e-01} (1.081e-05) & \cellcolor{gray!30}\textbf{7.340e-01} (8.937e-06) & \cellcolor{gray!30}\textbf{7.340e-01} (7.568e-05) \\
 & $(x_2)$ & \cellcolor{gray!30}\textbf{7.853e-01} (3.609e-05) & 7.852e-01 (3.939e-05) & \cellcolor{gray!30}\textbf{7.853e-01} (6.358e-05) \\
 & $(x_3)$ & \cellcolor{gray!30}\textbf{6.771e-01} (1.128e-04) & 6.769e-01 (9.895e-05) & \cellcolor{gray!30}\textbf{6.771e-01} (5.072e-05) \\
 & $(x_4)$ & \cellcolor{gray!30}\textbf{8.025e-01} (4.404e-05) & \cellcolor{gray!30}\textbf{8.025e-01} (3.540e-05) & \cellcolor{gray!30}\textbf{8.025e-01} (2.579e-05) \\
 & $(x_1,x_2)$ & \cellcolor{gray!30}\textbf{6.235e-01} (8.294e-06) & \cellcolor{gray!30}\textbf{6.235e-01} (1.242e-05) & \cellcolor{gray!30}\textbf{6.235e-01} (1.473e-05) \\
 & $(x_2,x_3)$ & \cellcolor{gray!30}\textbf{5.683e-01} (4.135e-05) & 5.682e-01 (2.144e-05) & 5.682e-01 (6.839e-05) \\
 & $(x_3,x_4)$ & 5.657e-01 (5.821e-05) & \cellcolor{gray!30}\textbf{5.658e-01} (3.339e-05) & \cellcolor{gray!30}\textbf{5.658e-01} (2.148e-05) \\
 & $(x_1,x_2,x_3)$ & \cellcolor{gray!30}\textbf{4.000e-01} (2.095e-05) & \cellcolor{gray!30}\textbf{4.000e-01} (1.627e-05) & \cellcolor{gray!30}\textbf{4.000e-01} (4.699e-06) \\
 & $(x_2,x_3,x_4)$ & \cellcolor{gray!30}\textbf{5.184e-01} (4.963e-05) & \cellcolor{gray!30}\textbf{5.184e-01} (2.300e-05) & \cellcolor{gray!30}\textbf{5.184e-01} (4.718e-05) \\
         \midrule
RE33 & $(x_1)$ & 8.468e-01 (8.575e-04) & \cellcolor{gray!30}\textbf{8.479e-01} (1.143e-03) & 8.444e-01 (4.743e-03) \\
 & $(x_2)$ & 8.086e-01 (1.850e-02) & 8.124e-01 (1.069e-02) & \cellcolor{gray!30}\textbf{8.189e-01} (3.444e-03) \\
 & $(x_3)$ & 5.301e-01 (5.034e-04) & \cellcolor{gray!30}\textbf{5.314e-01} (2.767e-04) & 4.963e-01 (5.024e-02) \\
 & $(x_4)$ & 8.287e-01 (1.110e-02) & \cellcolor{gray!30}\textbf{8.352e-01} (9.499e-03) & 8.282e-01 (7.405e-03) \\
 & $(x_1,x_2)$ & \cellcolor{gray!30}\textbf{4.921e-01} (9.278e-05) & 4.870e-01 (2.974e-03) & 4.867e-01 (1.064e-03) \\
 & $(x_2,x_3)$ & \cellcolor{gray!30}\textbf{4.672e-01} (3.151e-02) & 4.536e-01 (5.551e-02) & 4.546e-01 (5.296e-02) \\
 & $(x_3,x_4)$ & \cellcolor{gray!30}\textbf{3.879e-01} (2.744e-03) & 3.719e-01 (2.182e-02) & 3.870e-01 (1.130e-03) \\
 & $(x_1,x_2,x_3)$ & \cellcolor{gray!30}\textbf{2.212e-01} (7.541e-05) & \cellcolor{gray!30}\textbf{2.212e-01} (6.238e-05) & \cellcolor{gray!30}\textbf{2.211e-01} (7.949e-06) \\
 & $(x_2,x_3,x_4)$ & 7.672e-02 (1.692e-02) & \cellcolor{gray!30}\textbf{8.849e-02} (5.325e-05) & 8.833e-02 (4.638e-04) \\
    \bottomrule
    \end{tabular}
    \caption{Comparison of PPSL-MOBO between between rank size $r$.}
    \label{tab:hyper_sensitivity_r}
\end{table}

Finally, we assess the sensitivity to the LoRA rank $r$, which determines the capacity of the task-specific adapters. \Cref{tab:hyper_sensitivity_r} shows the performance for $r$ values of $\{2, 3, 5\}$. On the RE21 problem, the results are highly stable, with negligible performance differences across the tested ranks, indicating that a low-rank adaptation is sufficient to capture the parametric variations. For the more complex RE33 problem, the optimal rank varies with the shared component configuration; for instance, a smaller rank ($r=2$) is best for ($x_1, x_2$), while a larger rank ($r=5$) excels for ($x_2$). This suggests a trade-off between model expressiveness and the risk of overfitting. Overall, our default choice of $r=3$ (marked with an asterisk) strikes an effective balance, providing strong and reliable performance across the majority of configurations without introducing unnecessary model complexity.

\section{Experimental Analysis: Dynamic Multi-Objective Optimization}\label{app:sec:experiment_dmop}

\subsection{Experimental Setups}\label{subsec:dmop_experimental_set}

\subsubsection{Optimization Problems.}

To comprehensively evaluate performance in dynamic settings, we adopt the DF (Dynamic Functions) test suite~\cite{jiang2018benchmark}. The DF test suite features diverse dynamic characteristics, including varying degrees of convexity, linkage, and objective correlation, providing a robust platform for assessing algorithm adaptability.

In the DF suite, the dynamic environment is characterized by the normalized time index $t \in [0, 1]$, which is updated as:
\begin{equation}
    t = \frac{1}{n_t} \left\lfloor \frac{\tau}{\tau_t} \right\rfloor,
\end{equation}
where $\tau$ is the generation counter, $n_t$ is the severity of change, and $\tau_t$ is the frequency. In our experiments, we follow the standard protocol: the decision variable dimension is set to 10, population size is 100, and the environment undergoes 20 changes with a change frequency $\tau_t = 2$, meaning only two generations are available for adaptation at each change point.

\subsubsection{Algorithmic Integration}

A crucial adaptation for applying PPSL-MOBO to dynamic multi-objective optimization lies in the treatment of the task parameter. To faithfully reflect the online nature of DMOPs and ensure strict causality, the task distribution \( P_{\bm{t}} \) in Phase 2 of \Cref{alg:PPSL-MOBO-final} is set to a degenerate distribution at each time step. That is, at every time slot, all sampled task parameters \( \{\bm{t}_k\}_{k=1}^{N_t} \) are identical and correspond exactly to the \emph{current normalized time index}. 

Specifically, we define the time parameter as \( t = \tau/T_{\mathrm{max}} \), where \( \tau \) is the current generation counter and \( T_{\mathrm{max}} \) is the total number of generations. At each time step, this index is treated as a \emph{discrete and sequentially increasing parameter}, and no information from future time steps is used during training. This strict online formulation guarantees that PPSL-MOBO only leverages data up to the present time, mirroring the requirements of real-world dynamic optimization where future environments are inherently unknown. 

To enhance the accuracy of Gaussian Process modeling in dynamic environments, we propose a sample reuse strategy that leverages historical data from previous time steps. While the parametric model provides immediate adaptation to new environments, the GP surrogate can benefit from samples collected at earlier time points, especially when the environmental changes are gradual or when certain regions of the objective landscape remain relatively stable. However, since the computational complexity of GP modeling scales as $\mathcal{O}(n^3)$ with the number of training samples, indiscriminate accumulation of historical data becomes computationally prohibitive. To balance model accuracy and computational efficiency, we empirically limit the maximum number of samples to 200. When this limit is reached, we employ a first-in-first-out (FIFO) strategy, discarding the oldest samples to maintain computational tractability while ensuring the GP model remains responsive to recent environmental changes. This adaptive sample management allows PPSL-MOBO to build more accurate surrogate models without sacrificing the real-time performance crucial for dynamic optimization scenarios.

To align with the evaluation protocol of DMOPs, we adapt the acquisition function to match how algorithm performance is assessed in DMOPs. Since dynamic optimization benchmarks evaluate algorithm performance independently at each time step, measuring how well the algorithm tracks the moving Pareto front at each environment, we modify the HVI calculation accordingly. Specifically, we only consider the non-dominated solutions obtained within the current generation rather than accumulating solutions across multiple time steps. This design choice ensures that our acquisition strategy directly optimizes for the performance metric used in dynamic optimization evaluation, the quality of the Pareto front approximation at each individual time step. By focusing the acquisition function on immediate performance rather than cumulative progress, PPSL-MOBO effectively adapts its search behavior to the fundamental nature of dynamic optimization, where the goal is to quickly converge to each new Pareto front as the environment changes.

\subsubsection{Baseline Methods.}
To rigorously evaluate the efficiency and effectiveness of the proposed PPSL-MOBO method in dynamic multi-objective optimization, we benchmark against several state-of-the-art dynamic optimization algorithms. Specifically, we consider: (1) Dynamic MOEAs, including DNSGA-II-A and DNSGA-II-B~\cite{deb2007dynamic}, which are two widely adopted variants of NSGA-II designed with different change-detection and re-initialization strategies for dynamic environments; (2) Surrogate-assisted algorithms tailored for expensive DMOPs, including TL-MOEA/D-EGO~\cite{Fan2020Surrogate}, a transfer learning-based approach, and TrSA-DMOEA~\cite{zhang2023adaptive}, an advanced manifold transfer learning method leveraging geodesic flow kernels and knee solutions. All baseline implementations are based on the \texttt{pymoo} platform, with static modules and hyperparameters configured according to their respective original publications. The PPSL-MOBO method is implemented as described in Section~\ref{sec:method} and our open-source repository.

\subsubsection{Performance Indicators}

Algorithm performance is assessed using two standard dynamic metrics~\cite{zhou2013population, jiang2022evolutionary}:
\begin{itemize}
    \item Mean Inverted Generational Distance (MIGD): MIGD is a commonly used performance metric that evaluates both convergence and diversity. A smaller MIGD value indicates better algorithm performance. Let ${PF}_{t}^{*}, {PF}_{t}$ denote the true and approximate Pareto fronts respectively, then at time $t$ IGD can be calculated as:
\begin{equation}
    \mathrm{IGD}({PF}_{t}^{*},{PF}_{t})=\frac{1}{|{PF}_{t}^{*}|}\sum_{\bm{x}^{*}\in{PF}_{t}^{*}}\min_{\bm{x}\in{PF}_{t}}\|\bm{x}^{*}-\bm{x}\|_{2},
\end{equation}
and MIGD follows as:
\begin{equation}
    \mathrm{MIGD}=\frac{1}{|{T}|}\sum_{t\in{T}}\mathrm{IGD}({PF}_{t}^{*},{PF}_{t}),
\end{equation}
where $T$ denotes the time series and $|T|$ means the cardinality of $T$.
    \item Mean Hypervolume (MHV): MHV measures the mean hypervolume covered by the approximation. The MHV is defined as: 
\begin{equation}
    \operatorname{MHV}=\tfrac{1}{|{T}|}\sum_{t\in{T}}\mathrm{HV}({PF}_{t}\mid z_{\text{ref}}(t))
\end{equation}
, where the reference vector is $z_{\text{ref}}(t)=(z_{1}(t),\dots ,z_{m}(t))$ and $z_{i}(t)=\max_{\bm{y}\in{PF}_{t}^{*}}y_{i}$.
\end{itemize}

For PPSL-MOBO, we report results when generating 200 solutions per time step. All experiments are repeated for 10 independent runs, and the mean and standard deviations of MIGD and MHV are reported.

\begin{table}
    \centering
    \scriptsize
    \begin{tabular}{cc|ccccc}
        \toprule[1.0pt]
        Problems & $n_t$ & DNSGA-II-A & DNSGA-II-B &  TL-MOEA/D-EGO* & TrSA-DMOEA* & PPSL-MOBO \\
        \midrule 
DF1 & 5 & 3.634e-01 (3.556e-02) & 6.329e-01 (9.949e-02) & 1.980e-01 (1.270e-01) & 2.989e-01 (7.100e-02) & \cellcolor{gray!30}\textbf{5.269e-02} (2.332e-03) \\
& 10 & 3.318e-01 (1.402e-02) & 6.030e-01 (1.091e-02) & 1.682e-01 (6.200e-02) & 2.654e-01 (6.100e-02) & \cellcolor{gray!30}\textbf{6.172e-02} (1.628e-02) \\
& 20 & 2.742e-01 (1.403e-02) & 5.125e-01 (4.538e-02) & 1.232e-01 (7.200e-02) & 2.749e-01 (5.900e-02) & \cellcolor{gray!30}\textbf{7.499e-02} (1.880e-02) \\
        \midrule
DF2 & 5 & 2.412e-01 (9.748e-03) & 3.998e-01 (2.802e-02) & 1.305e-01 (6.900e-02) & \cellcolor{gray!30}\textbf{1.660e-02} (9.000e-03) & 1.032e-01 (3.805e-02) \\
& 10 & 1.890e-01 (5.011e-03) & 3.818e-01 (1.093e-02) & 1.225e-01 (6.200e-02) & \cellcolor{gray!30}\textbf{1.250e-02} (7.000e-03) & 1.310e-01 (1.918e-02) \\
& 20 & 1.666e-01 (1.099e-02) & 3.737e-01 (2.401e-02) & 7.720e-02 (4.000e-02) & \cellcolor{gray!30}\textbf{1.180e-02} (6.000e-03) & 1.189e-01 (1.899e-02) \\
        \midrule
DF3 & 5 & 8.529e-01 (7.640e-02) & 2.234e+00 (2.950e-01) & 7.183e-01 (2.810e-01) & \cellcolor{gray!30}\textbf{5.150e-02} (2.500e-02) & 2.939e-01 (1.656e-01) \\
& 10 & 9.689e-01 (1.379e-02) & 8.750e-01 (1.797e-01) & 6.122e-01 (2.440e-01) & 5.020e-02 (2.700e-02) & \cellcolor{gray!30}\textbf{4.523e-02} (1.598e-02) \\
& 20 & 9.007e-01 (4.061e-02) & 8.605e-01 (7.784e-02) & 6.640e-01 (2.980e-01) & 5.820e-02 (3.100e-02) & \cellcolor{gray!30}\textbf{2.618e-02} (1.318e-02) \\
        \midrule
DF4 & 5 & 1.581e+00 (2.395e-01) & 1.030e+00 (1.552e-01) & 2.263e+00 (6.880e-01) & 1.228e+00 (4.230e-01) & \cellcolor{gray!30}\textbf{1.918e-01} (5.183e-03) \\
& 10 & 1.262e+00 (2.101e-01) & 1.042e+00 (3.826e-02) & 2.086e+00 (6.900e-01) & 1.236e+00 (4.120e-01) & \cellcolor{gray!30}\textbf{1.188e-01} (1.224e-02) \\
& 20 & 1.369e+00 (2.987e-01) & 1.448e+00 (2.384e-01) & 2.421e+00 (7.110e-01) & 1.240e+00 (4.140e-01) & \cellcolor{gray!30}\textbf{8.505e-02} (1.577e-03) \\
        \midrule
DF5 & 5 & 7.596e-01 (3.739e-02) & 3.284e+00 (3.655e-01) & 1.687e-01 (1.290e-01) & \cellcolor{gray!30}\textbf{1.710e-02} (1.400e-02) & 2.910e-02 (1.018e-02) \\
& 10 & 4.820e-01 (3.880e-02) & 4.950e-01 (2.807e-02) & 1.226e-01 (1.050e-01) & 1.840e-02 (1.300e-02) & \cellcolor{gray!30}\textbf{1.773e-02} (3.979e-03) \\
& 20 & 3.765e-01 (3.077e-02) & 3.607e-01 (1.296e-02) & 1.034e-01 (6.400e-02) & 1.790e-02 (1.100e-02) & \cellcolor{gray!30}\textbf{6.849e-03} (8.349e-04) \\
        \midrule
DF6 & 5 & 1.433e+01 (7.229e-01) & 1.926e+01 (3.744e-01) & 3.524e+00 (2.738e+00) & \cellcolor{gray!30}\textbf{2.068e+00} (1.926e+00) & 1.594e+01 (9.084e-01) \\
& 10 & 1.281e+01 (3.672e-01) & 1.809e+01 (1.527e+00) & 3.140e+00 (2.508e+00) & \cellcolor{gray!30}\textbf{1.892e+00} (1.193e+00) & 3.396e+01 (3.210e+01) \\
& 20 & 1.144e+01 (6.410e-01) & 1.305e+01 (9.611e-01) & 3.113e+00 (2.856e+00) & \cellcolor{gray!30}\textbf{1.756e+00} (1.210e+00) & 3.408e+01 (3.201e+01) \\
        \midrule
DF7 & 5 & 4.814e-01 (8.200e-03) & 5.201e-01 (1.436e-02) & 3.863e+00 (4.364e+00) & 2.221e+00 (1.629e+00) & \cellcolor{gray!30}\textbf{1.585e-02} (2.281e-04) \\
& 10 & 4.450e-01 (3.253e-02) & 4.125e-01 (2.247e-02) & 3.581e+00 (3.974e+00) & 2.112e+00 (1.827e+00) & \cellcolor{gray!30}\textbf{1.142e-02} (7.530e-04) \\
& 20 & 3.866e-01 (2.507e-02) & 3.765e-01 (2.441e-02) & 3.412e+00 (3.914e+00) & 2.094e+00 (1.928e+00) & \cellcolor{gray!30}\textbf{1.034e-02} (1.780e-03) \\
        \midrule
DF8 & 5 & 2.026e-01 (6.161e-03) & 1.870e-01 (1.900e-02) & 9.859e-01 (4.690e-01) & 3.732e-01 (7.100e-02) & \cellcolor{gray!30}\textbf{3.913e-02} (5.410e-03) \\
& 10 & 2.019e-01 (2.456e-02) & 1.783e-01 (2.142e-02) & 8.573e-01 (4.130e-01) & 2.839e-01 (6.800e-02) & \cellcolor{gray!30}\textbf{4.286e-02} (8.838e-03) \\
& 20 & 1.981e-01 (3.533e-03) & 1.565e-01 (1.357e-02) & 9.088e-01 (4.240e-01) & 1.794e-01 (4.100e-02) & \cellcolor{gray!30}\textbf{4.908e-02} (7.736e-03) \\
        \midrule
DF9 & 5 & \cellcolor{gray!30}\textbf{1.310e+00} (4.752e-02) & 1.802e+00 (7.227e-02) & 2.040e+00 (2.599e+00) & 2.222e+00 (2.310e+00) & 1.821e+00 (4.884e-01) \\
& 10 & 1.041e+00 (1.161e-01) & 2.142e+00 (4.726e-01) & 2.026e+00 (2.471e+00) & 2.218e+00 (1.914e+00) & \cellcolor{gray!30}\textbf{1.053e+00} (1.015e-01) \\
& 20 & 6.554e-01 (8.274e-02) & 8.668e-01 (1.726e-01) & 2.082e+00 (2.363e+00) & 2.173e+00 (2.117e+00) & \cellcolor{gray!30}\textbf{1.513e-01} (4.871e-02) \\
        \midrule
DF10 & 5 & 4.188e-01 (4.198e-02) & 3.331e-01 (2.353e-02) & 2.761e-01 (1.570e-01) & \cellcolor{gray!30}\textbf{1.266e-01} (2.100e-02) & 2.167e-01 (4.159e-02) \\
& 10 & 3.664e-01 (1.154e-02) & 3.040e-01 (2.164e-02) & 2.581e-01 (1.390e-01) & \cellcolor{gray!30}\textbf{1.217e-01} (2.000e-02) & 1.648e-01 (1.497e-02) \\
& 20 & 2.913e-01 (9.813e-03) & 1.929e-01 (1.768e-02) & 2.726e-01 (1.530e-01) & 1.663e-01 (2.600e-02) & \cellcolor{gray!30}\textbf{8.481e-02} (1.151e-02) \\
        \midrule
DF11 & 5 & 1.169e+01 (4.684e-02) & 1.170e+01 (6.148e-02) & 2.839e-01 (8.200e-02) & \cellcolor{gray!30}\textbf{1.699e-01} (4.800e-02) & 3.174e-01 (1.822e-03) \\
& 10 & 1.155e+01 (7.668e-02) & 1.161e+01 (3.951e-02) & 2.673e-01 (6.200e-02) & \cellcolor{gray!30}\textbf{1.653e-01} (4.300e-02) & 3.748e-01 (7.965e-02) \\
& 20 & 1.148e+01 (1.073e-02) & 1.160e+01 (1.356e-02) & 2.568e-01 (6.800e-02) & \cellcolor{gray!30}\textbf{1.569e-01} (4.200e-02) & 3.150e-01 (6.504e-03) \\
        \midrule
DF12 & 5 & 6.591e-01 (7.776e-02) & 5.511e-01 (2.718e-02) & 1.198e+00 (2.940e-01) & 3.396e-01 (9.100e-02) & \cellcolor{gray!30}\textbf{1.021e-01} (1.329e-02) \\
& 10 & 6.184e-01 (4.629e-02) & 5.494e-01 (1.566e-02) & 1.137e+00 (2.830e-01) & 3.221e-01 (1.110e-01) & \cellcolor{gray!30}\textbf{6.742e-02} (5.661e-03) \\
& 20 & 4.983e-01 (3.962e-02) & 4.191e-01 (1.585e-02) & 1.078e+00 (2.890e-01) & 3.448e-01 (1.420e-01) & \cellcolor{gray!30}\textbf{4.815e-02} (4.864e-03) \\
        \midrule
DF13 & 5 & 1.303e+00 (9.085e-02) & 2.282e+00 (5.165e-01) & 1.512e+00 (1.460e+00) & 5.126e-01 (3.420e-01) & \cellcolor{gray!30}\textbf{9.850e-02} (9.124e-03) \\
& 10 & 8.193e-01 (7.735e-02) & 7.307e-01 (7.456e-02) & 1.443e+00 (1.193e+00) & 4.356e-01 (2.280e-01) & \cellcolor{gray!30}\textbf{8.403e-02} (6.590e-03) \\
& 20 & 7.315e-01 (6.209e-02) & 6.228e-01 (5.053e-02) & 1.377e+00 (1.334e+00) & 4.158e-01 (3.210e-01) & \cellcolor{gray!30}\textbf{6.968e-02} (1.865e-03) \\
        \midrule
DF14 & 5 & 8.095e-01 (1.055e-01) & 2.390e+00 (2.265e-01) & 9.292e-01 (5.510e-01) & 2.036e-01 (1.330e-01) & \cellcolor{gray!30}\textbf{4.069e-02} (6.082e-03) \\
& 10 & 3.954e-01 (1.950e-02) & 3.768e-01 (9.131e-03) & 8.872e-01 (4.780e-01) & 1.959e-01 (1.150e-01) & \cellcolor{gray!30}\textbf{3.226e-02} (7.868e-04) \\
& 20 & 3.607e-01 (2.999e-02) & 3.300e-01 (3.792e-02) & 8.901e-01 (5.390e-01) & 2.544e-01 (1.020e-01) & \cellcolor{gray!30}\textbf{2.505e-01} (1.602e-01) \\
        \bottomrule[1.0pt]
    \end{tabular}
    \caption{The average and standard deviation MIGD values of ten independent runs on benchmark problems with three different dynamic environment. The best results in each test are highlighted in bold and gray background. The results for TL-MOEA/D-EGO* and TrSA-DMOEA* are directly taken from \cite{zhang2023adaptive}.}
    \label{tab:dmop_migd}
\end{table}

\subsection{Experimental Results}

Table \ref{tab:dmop_migd} presents the MIGD performance comparison across 14 DF benchmark problems under three different change severities ($n_t \in \{5, 10, 20\}$). The results reveal several key insights:

\subsubsection{Overall Performance. } PPSL-MOBO demonstrates superior performance on the majority of test problems, achieving the best MIGD values in 24 out of 42 test instances. This strong performance is particularly notable given that PPSL-MOBO operates under the strict online constraint, using only historical data up to the current time step, while maintaining competitive computational efficiency through our sample reuse strategy.
The experimental results validate that parametric learning of Pareto sets provides an effective paradigm for expensive dynamic multi-objective optimization, offering immediate adaptation to environmental changes while maintaining high solution quality across diverse problem characteristics.

\subsubsection{Performance vs. Change Severity. }
As the severity of environmental changes increases (larger $n_t$), PPSL-MOBO shows remarkable robustness. For instance, on DF1, DF4, and DF7, PPSL-MOBO consistently outperforms all baselines across all severity levels, with performance gaps widening as $n_t$ increases. This suggests that the parametric learning approach becomes increasingly advantageous when environmental changes are more dramatic.

\subsubsection{Comparison with Surrogate-assisted Methods.}
When compared against state-of-the-art surrogate-assisted methods (TL-MOEA/D-EGO and TrSA-DMOEA), PPSL-MOBO shows competitive or superior performance on most problems. While TrSA-DMOEA excels on certain problems like DF2, DF3, and DF5, PPSL-MOBO achieves significantly better results on DF1, DF4, DF7, DF8, DF12, DF13, and DF14. This indicates that the parametric approach offers complementary strengths to transfer learning-based methods.

\subsubsection{Challenging Scenarios.}
PPSL-MOBO encounters difficulties on DF6 and DF10, where transfer learning methods perform better. These problems likely contain characteristics that favor the manifold transfer learning approach of TrSA-DMOEA. However, even in these cases, PPSL-MOBO maintains reasonable performance and often outperforms the traditional dynamic MOEAs.

\begin{figure}
    \centering
    \includegraphics[width=0.9\linewidth]{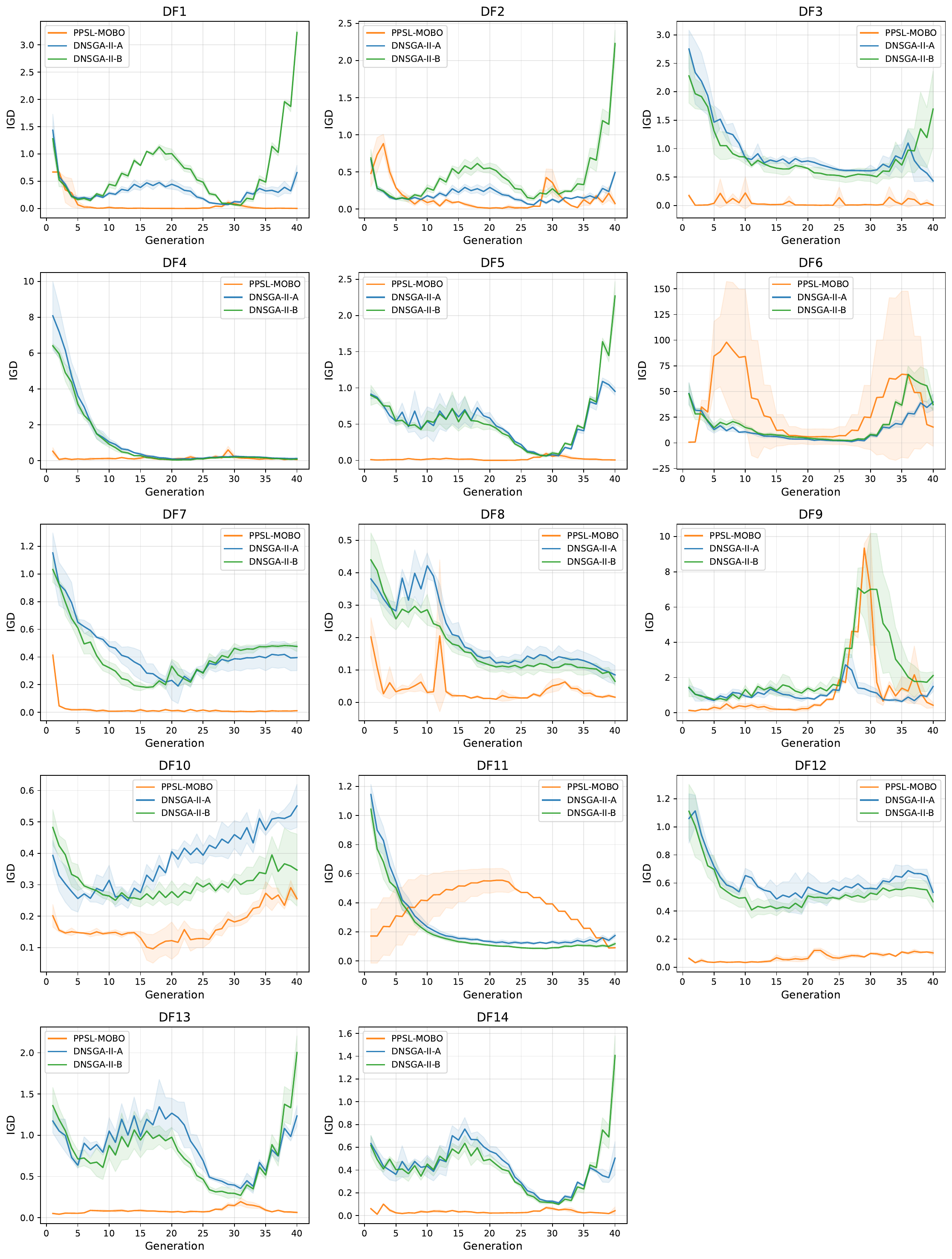}
    \caption{Evolution of MIGD performance over time for PPSL-MOBO and DNSGA-II variants across 14 DF benchmark problems with change severity $n_t = 10$. The $x$-axis represents generations (time steps), and the $y$-axis shows IGD values (lower is better). Environmental changes occur every 2 generations, creating the characteristic periodic patterns in algorithm performance.}
    \label{fig:dmop_migd}
\end{figure}

\begin{figure}
    \centering
    \includegraphics[width=0.9\linewidth]{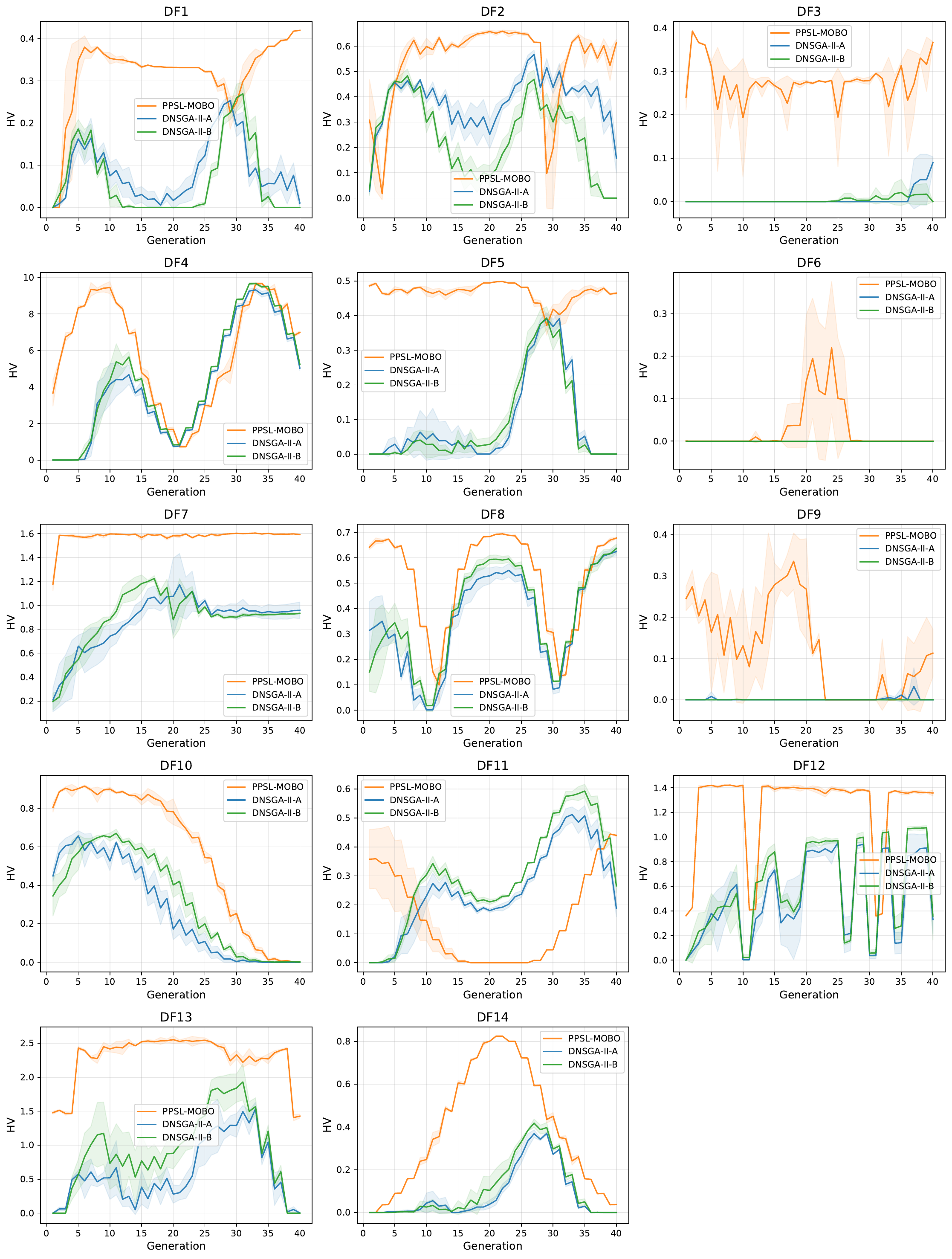}
    \caption{Evolution of MHV performance over time for PPSL-MOBO and DNSGA-II variants across 14 DF benchmark problems with change severity $n_t = 10$. The $x$-axis represents generations (time steps), and the $y$-axis shows HV values (higher is better). Environmental changes occur every 2 generations.}
    \label{fig:dmop_mhv}
\end{figure}

\begin{table}
    \centering
    \scriptsize
    \begin{tabular}{cc|cccc}
        \toprule[1.0pt]
        Problems & $n_t$ & PPSL-MOBO-NoHistory & PPSL-MOBO-NN &  PPSL-MOBO-Random & PPSL-MOBO \\
        \midrule 
DF1 & 5 & 1.309e-01 (8.447e-02) & 3.741e-01 (4.447e-02) & 1.622e-01 (1.280e-02) & \cellcolor{gray!30}\textbf{5.269e-02} (2.332e-03) \\
& 10 & 1.038e-01 (3.035e-02) & 2.232e-01 (4.184e-02) & 2.601e-01 (1.452e-02) & \cellcolor{gray!30}\textbf{6.172e-02} (1.628e-02) \\
& 20 & 1.896e-01 (1.200e-01) & 2.017e-01 (2.798e-02) & 5.054e-01 (4.035e-03) & \cellcolor{gray!30}\textbf{7.499e-02} (1.880e-02) \\
        \midrule
DF2 & 5 & 1.264e-01 (2.738e-02) & 4.169e-01 (7.516e-02) & 1.477e-01 (3.497e-02) & \cellcolor{gray!30}\textbf{1.032e-01} (3.805e-02) \\
& 10 & \cellcolor{gray!30}\textbf{1.231e-01} (1.305e-02) & 2.741e-01 (6.377e-03) & 1.362e-01 (7.363e-03) & 1.310e-01 (1.918e-02) \\
& 20 & 1.801e-01 (2.551e-02) & 2.630e-01 (4.143e-02) & 3.117e-01 (3.549e-03) & \cellcolor{gray!30}\textbf{1.189e-01} (1.899e-02) \\
        \midrule
DF3 & 5 & 8.482e-01 (1.305e-01) & 1.191e+00 (1.948e-01) & 4.037e-01 (3.323e-02) & \cellcolor{gray!30}\textbf{2.939e-01} (1.656e-01) \\
& 10 & 5.856e-01 (1.578e-02) & 1.408e+00 (3.571e-02) & 3.998e-01 (5.786e-02) & \cellcolor{gray!30}\textbf{4.523e-02} (1.598e-02) \\
& 20 & 5.884e-01 (1.186e-02) & 1.288e+00 (1.667e-01) & 9.678e-01 (2.770e-02) & \cellcolor{gray!30}\textbf{2.618e-02} (1.318e-02) \\
        \midrule
DF4 & 5 & 4.075e-01 (5.745e-03) & 2.932e+00 (4.853e-01) & 1.455e+00 (3.469e-01) & \cellcolor{gray!30}\textbf{1.918e-01} (5.183e-03) \\
& 10 & 4.135e-01 (2.497e-02) & 3.447e+00 (4.218e-01) & 1.619e+00 (1.137e-01) & \cellcolor{gray!30}\textbf{1.188e-01} (1.224e-02) \\
& 20 & 3.507e-01 (2.904e-02) & 2.693e+00 (1.694e-01) & 3.326e+00 (5.095e-02) & \cellcolor{gray!30}\textbf{8.505e-02} (1.577e-03) \\
        \midrule
DF5 & 5 & 6.076e-02 (1.838e-02) & 6.879e-01 (1.419e-01) & 3.637e-01 (4.743e-02) & \cellcolor{gray!30}\textbf{2.910e-02} (1.018e-02) \\
& 10 & 3.726e-02 (5.734e-03) & 5.235e-01 (9.362e-02) & 2.161e-01 (3.339e-02) & \cellcolor{gray!30}\textbf{1.773e-02} (3.979e-03) \\
& 20 & 1.619e-02 (2.306e-03) & 4.069e-01 (5.593e-03) & 4.109e-01 (3.600e-02) & \cellcolor{gray!30}\textbf{6.849e-03} (8.349e-04) \\
        \midrule
DF6 & 5 & 2.207e+01 (1.688e+01) & 4.697e+01 (2.411e+01) & 2.404e+01 (1.703e+00) & \cellcolor{gray!30}\textbf{1.594e+01} (9.084e-01) \\
& 10 & \cellcolor{gray!30}\textbf{3.286e+00} (1.834e+00) & 2.166e+01 (3.994e-01) & 2.063e+01 (1.198e+00) & 3.396e+01 (3.210e+01) \\
& 20 & \cellcolor{gray!30}\textbf{7.225e-01} (1.148e-01) & 4.652e+01 (2.519e+01) & 5.962e+01 (1.007e+01) & 3.408e+01 (3.201e+01) \\
        \midrule
DF7 & 5 & 1.281e-01 (2.769e-02) & 4.863e-01 (1.086e-01) & 7.022e-01 (3.324e-01) & \cellcolor{gray!30}\textbf{1.585e-02} (2.281e-04) \\
& 10 & 1.478e-01 (3.389e-02) & 3.640e-01 (6.851e-02) & 4.262e-01 (7.986e-02) & \cellcolor{gray!30}\textbf{1.142e-02} (7.530e-04) \\
& 20 & 1.460e-01 (3.632e-02) & 3.010e-01 (1.037e-01) & 9.004e-01 (4.499e-03) & \cellcolor{gray!30}\textbf{1.034e-02} (1.780e-03) \\
        \midrule
DF8 & 5 & 5.428e-02 (7.686e-03) & 3.105e-01 (1.391e-02) & 1.121e-01 (2.134e-02) & \cellcolor{gray!30}\textbf{3.913e-02} (5.410e-03) \\
& 10 & 5.943e-02 (1.129e-02) & 2.277e-01 (4.080e-03) & 1.514e-01 (2.911e-02) & \cellcolor{gray!30}\textbf{4.286e-02} (8.838e-03) \\
& 20 & 6.819e-02 (2.273e-02) & 2.065e-01 (3.895e-02) & 2.468e-01 (4.665e-02) & \cellcolor{gray!30}\textbf{4.908e-02} (7.736e-03) \\
        \midrule
DF9 & 5 & 2.998e+00 (2.626e+00) & 1.875e+00 (2.482e-01) & \cellcolor{gray!30}\textbf{1.188e+00} (1.109e-01) & 1.821e+00 (4.884e-01) \\
& 10 & 1.815e+00 (4.640e-01) & 1.146e+00 (2.189e-01) & 1.416e+00 (3.913e-01) & \cellcolor{gray!30}\textbf{1.053e+00} (1.015e-01) \\
& 20 & 4.137e-01 (1.144e-01) & 9.738e-01 (8.225e-02) & 1.568e+00 (1.672e-02) & \cellcolor{gray!30}\textbf{1.513e-01} (4.871e-02) \\
        \midrule
DF10 & 5 & 4.308e-01 (5.618e-02) & 4.266e-01 (1.478e-02) & 3.248e-01 (3.893e-02) & \cellcolor{gray!30}\textbf{2.167e-01} (4.159e-02) \\
& 10 & 3.575e-01 (4.486e-02) & 3.765e-01 (1.214e-02) & 2.556e-01 (6.846e-03) & \cellcolor{gray!30}\textbf{1.648e-01} (1.497e-02) \\
& 20 & 1.655e-01 (5.470e-03) & 2.169e-01 (1.302e-02) & 1.943e-01 (5.312e-03) & \cellcolor{gray!30}\textbf{8.481e-02} (1.151e-02) \\
        \midrule
DF11 & 5 & 3.277e-01 (3.697e-03) & 4.750e-01 (4.136e-02) & 5.955e-01 (5.140e-02) & \cellcolor{gray!30}\textbf{3.174e-01} (1.822e-03) \\
& 10 & \cellcolor{gray!30}\textbf{3.311e-01} (8.960e-04) & 3.816e-01 (2.327e-02) & 7.008e-01 (2.139e-02) & 3.748e-01 (7.965e-02) \\
& 20 & 3.436e-01 (1.146e-02) & 3.380e-01 (2.874e-03) & 7.541e-01 (2.180e-02) & \cellcolor{gray!30}\textbf{3.150e-01} (6.504e-03) \\
        \midrule
DF12 & 5 & 1.755e-01 (4.344e-02) & 6.944e-01 (8.420e-02) & 5.795e-01 (2.274e-01) & \cellcolor{gray!30}\textbf{1.021e-01} (1.329e-02) \\
& 10 & 1.111e-01 (1.906e-02) & 4.702e-01 (2.506e-02) & 3.385e-01 (4.834e-02) & \cellcolor{gray!30}\textbf{6.742e-02} (5.661e-03) \\
& 20 & 8.487e-02 (1.495e-03) & 4.193e-01 (1.134e-01) & 2.170e-01 (4.355e-03) & \cellcolor{gray!30}\textbf{4.815e-02} (4.864e-03) \\
        \midrule
DF13 & 5 & 2.323e-01 (4.974e-02) & 3.669e-01 (2.288e-02) & 5.762e-01 (5.851e-02) & \cellcolor{gray!30}\textbf{9.850e-02} (9.124e-03) \\
& 10 & 2.294e-01 (1.702e-02) & 3.513e-01 (9.296e-03) & 4.929e-01 (5.552e-02) & \cellcolor{gray!30}\textbf{8.403e-02} (6.590e-03) \\
& 20 & 1.369e-01 (3.620e-02) & 3.061e-01 (3.742e-02) & 7.727e-01 (3.004e-02) & \cellcolor{gray!30}\textbf{6.968e-02} (1.865e-03) \\
        \midrule
DF14 & 5 & 6.986e-02 (1.112e-02) & 4.540e-01 (3.787e-02) & 5.009e-01 (4.836e-02) & \cellcolor{gray!30}\textbf{4.069e-02} (6.082e-03) \\
& 10 & 1.034e-01 (3.882e-02) & 2.205e-01 (1.902e-02) & 3.381e-01 (3.527e-02) & \cellcolor{gray!30}\textbf{3.226e-02} (7.868e-04) \\
& 20 & 3.726e-01 (2.938e-04) & \cellcolor{gray!30}\textbf{1.532e-01} (7.306e-03) & 4.626e-01 (1.090e-02) & 2.605e-01 (1.602e-01) \\
        \bottomrule[1.0pt]
    \end{tabular}
    \caption{The average and standard deviation MIGD values for the PPSL-MOBO variants.}
    \label{tab:dmop_migd_abla}
\end{table}

\subsubsection{Temporal Performance Analysis. }
\Cref{fig:dmop_migd} and \Cref{fig:dmop_mhv} provide complementary views of algorithm behavior through MIGD and MHV metrics, revealing consistent patterns in dynamic adaptation. The MHV results corroborate the findings from MIGD analysis while offering additional insights into solution diversity:

\textit{Convergence-Diversity Trade-off:} While MIGD primarily captures convergence quality, MHV reflects both convergence and diversity. On problems like DF1 and DF2, PPSL-MOBO achieves superior MHV performance, indicating that it not only converges closer to the true Pareto front but also maintains better coverage. The DNSGA-II variants show characteristic drops in MHV at each environmental change, followed by gradual recovery, a pattern that mirrors their MIGD behavior but is often more pronounced due to diversity loss during reinitialization.

\textit{Stability vs. Adaptability:} PPSL-MOBO demonstrates remarkable stability in both metrics across most problems, maintaining consistent performance without the dramatic fluctuations observed in evolutionary approaches. This is particularly evident in DF4, DF7, and DF12, where PPSL-MOBO's MHV curves remain nearly flat while DNSGA-II variants exhibit significant oscillations. However, on problems like DF3 and DF6, PPSL-MOBO shows delayed response in MHV, suggesting that while the parametric model provides stable solutions, it may occasionally struggle to capture the full extent of Pareto front changes in certain complex landscapes.

\textit{Early vs. Late Performance:} An interesting pattern emerges in problems DF6 and DF11, where PPSL-MOBO initially outperforms DNSGA-II variants but gradually loses its advantage. This suggests that for certain problem characteristics, the evolutionary algorithms' ability to accumulate population information over time can eventually overcome the immediate adaptation advantage of parametric learning. Nevertheless, PPSL-MOBO's consistent early performance makes it particularly valuable in scenarios requiring quick response to environmental changes or when computational budgets are severely limited.

\subsection{Ablation Study}

To investigate the contribution of each key component in PPSL-MOBO for dynamic optimization, we conduct an ablation study with three variants:
\begin{itemize}
    \item PPSL-MOBO-NoHistory: This variant builds GP surrogate models using only samples from the current time step, excluding historical data from previous environments. By removing the temporal information transfer, we assess the importance of leveraging past experiences in adapting to environmental changes.
    \item PPSL-MOBO-NN: In this variant, we replace the Gaussian Process models with deterministic neural networks (NNs). Since NNs cannot provide uncertainty estimates like GPs, this variant essentially becomes a greedy approach that directly learns the mapping without uncertainty-guided exploration. This allows us to evaluate the role of uncertainty quantification in balancing exploration and exploitation.
    \item PPSL-MOBO-Random: This variant removes the intelligent acquisition function and instead randomly samples solutions at each generation. By comparing with this baseline, we can quantify the effectiveness of our uncertainty-based acquisition strategy in guiding the search process toward promising regions of the dynamic Pareto front.
\end{itemize}

These ablation studies help decompose the performance gains of PPSL-MOBO and identify which components are most critical for effective dynamic multi-objective optimization under expensive evaluation settings. The results for MIGD indicator are presented in \Cref{tab:dmop_migd_abla}.

\subsubsection{Impact of Historical Information (PPSL-MOBO vs. PPSL-MOBO-NoHistory):} The comparison demonstrates the significant value of incorporating temporal information in GP modeling. PPSL-MOBO consistently outperforms NoHistory across nearly all problems, with particularly dramatic improvements on DF3, DF5, DF7, etc. This validates our adaptive GP modeling strategy that leverages samples from previous time steps to capture environmental correlations. The performance gap is especially pronounced in problems with strong temporal dependencies, where historical information provides crucial context for predicting the evolved Pareto front. Interestingly, NoHistory shows competitive performance only on DF6, suggesting that for certain highly volatile environments, relying too heavily on past information may occasionally be detrimental.

\subsubsection{Role of Uncertainty Quantification (PPSL-MOBO vs. PPSL-MOBO-NN):} Replacing GP models with deterministic neural networks leads to substantial performance degradation across all test problems. The NN variant consistently produces MIGD values 2-10 times worse than PPSL-MOBO, with particularly poor performance on DF4 and DF6. This stark difference underscores the critical role of uncertainty estimates in guiding efficient exploration. Without uncertainty quantification, the greedy NN approach fails to balance exploration and exploitation effectively, leading to premature convergence to suboptimal regions. The consistent underperformance across all problem types confirms that uncertainty-guided search is essential for sample-efficient optimization in expensive settings.

\subsubsection{Effectiveness of Acquisition Strategy (PPSL-MOBO vs. PPSL-MOBO-Random):} The random sampling variant performs significantly worse than PPSL-MOBO in most cases, validating the importance of our intelligent acquisition function. The performance gap is particularly evident in problems requiring focused exploration, such as DF7 and DF13. However, Random shows surprisingly competitive results on DF9 at $n_t=5$, suggesting that for certain problem landscapes, especially in early stages, broad exploration through random sampling can occasionally be beneficial. Nevertheless, the overall superior performance of PPSL-MOBO confirms that our uncertainty-based acquisition strategy effectively identifies promising regions for evaluation, making optimal use of the limited evaluation budget.

The ablation results collectively demonstrate that all three components, Temporal Information Transfer, Uncertainty Quantification, and Intelligent Acquisition, work synergistically to achieve PPSL-MOBO's strong performance in expensive dynamic multiobjective optimization.

\section{Licenses}\label{app:sec:license}

\Cref{tab:licenses} lists the licenses for the codes and problem suite we used in this work.

\begin{table}[h]
    \scriptsize
    \centering
    \begin{tabular}{llll}
    \toprule[1.5pt]
    Resource & Type & Link & License \\
    \midrule
        BoTorch \cite{balandat2020botorch} & Code & \url{https://botorch.org/} & MIT License \\
        pymoo \cite{blank2020pymoo} & Code & \url{https://pymoo.org/} & Apache License 2.0 \\ 
        RE problems \cite{tanabe2020easy} & Problem Suite & \url{https://ryojitanabe.github.io/reproblems/} & None \\ 
        PSL-MOBO \cite{lin2022paretonips} & Code & \url{https://github.com/Xi-L/PSL-MOBO} & MIT License \\ 
        PPSL-MOBO (Ours) & Code & \url{https://github.com/jicheng9617/PPSL-MOBO} & MIT License \\
        \bottomrule[1.5pt]
    \end{tabular}
    \caption{List of licenses for the codes and problem suite we used in this work.}
    \label{tab:licenses}
\end{table}

\newpage

\ifdefstring{\appendicesmode}{true}{%
  \bibliography{ref_appendix}%
}{%
}

\end{document}